\documentclass[journal]{IEEEtran}

\ifCLASSINFOpdf

\else

\fi

\usepackage{soul}
\usepackage{cite}
\usepackage{amsmath,amssymb,amsfonts}
\usepackage{algorithmic}
\usepackage{graphicx}
\usepackage{textcomp}
\usepackage{xcolor}
\usepackage{booktabs}
\usepackage{multirow}
\usepackage{caption}
\usepackage{subcaption}
\usepackage[switch,pagewise]{lineno}
\usepackage{algorithm}
\usepackage{algorithmic}
\usepackage{amsmath}
\usepackage{bm}
\usepackage{threeparttable}
\usepackage[colorlinks,linkcolor=blue]{hyperref}
\usepackage{hyperref}
\usepackage{array}
\usepackage{colortbl}

\soulregister\cite7

\begin{document}

\title{PeQuENet: Perceptual Quality Enhancement of Compressed Video with Adaptation- and Attention-based Network}

\author{Saiping~Zhang,
         Luis~Herranz,
         Marta~Mrak,~\IEEEmembership{Senior Member,~IEEE,}
         Marc~G\'orriz~Blanch,
         Shuai~Wan
         and~Fuzheng~Yang
\thanks{S. Zhang and F. Yang are with the State Key Laboratory
of Integrated Services Networks, Xidian University, Xi’an, China e-mail: (spzhang@stu.xidian.edu.cn, fzhyang@mail.xidian.edu.cn).}
\thanks{L. Herranz is with the Computer Vision Center, Universitat Autònoma de Barcelona, 08193 Barcelona, Spain e-mail: (lherranz@cvc.uab.es).}
\thanks{M. Mrak and M. G. Blanch are with BBC Research \& Development, The Lighthouse, White City Place, 201 Wood Lane, London, UK (e-mail: marta.mrak@bbc.co.uk, marc.gorrizblanch@bbc.co.uk).}
\thanks{S. Wan is with the School of Electronics and Information, Northwestern Polytechnical University, Xi’an, China e-mail: (swan@nwpu.edu.cn).}}


\maketitle

\begin{abstract}
In this paper we propose a generative adversarial network (GAN) framework to enhance the perceptual quality of compressed videos. Our framework includes attention and adaptation to different quantization parameters (QPs) in a single model. The attention module exploits global receptive fields that can capture and align long-range correlations between consecutive frames, which can be beneficial for enhancing perceptual quality of videos. The frame to be enhanced is fed into the deep network together with its neighboring frames, and in the first stage features at different depths are extracted. Then extracted features are fed into attention blocks to explore global temporal correlations, followed by a series of upsampling and convolution layers. Finally, the resulting features are processed by the QP-conditional adaptation module which leverages the corresponding QP information. In this way, a single model can be used to enhance adaptively to various QPs without requiring multiple models specific for every QP value, while having similar performance. Experimental results demonstrate the superior performance of the proposed PeQuENet compared with the state-of-the-art compressed video quality enhancement algorithms.
\end{abstract}


\IEEEpeerreviewmaketitle

\section{Introduction}
\IEEEPARstart{R}{ecent} years have witnessed the rapid development in video services. Due to the limitation of bandwidth, video compression algorithms and video coding formats such as H.264/AVC \cite{b1} and H.265/HEVC \cite{b2} have been indispensable to remove the spatial and temporal redundancy in videos and reduce bit-rates. However, lossy video compression also degrades the quality of compressed videos and introduces various compression artifacts, such as blocking and blurring, which inevitably lead to the deterioration of the quality of experience (QoE) of the final user~\cite{b3}. 
Thus, effective algorithms for video quality enhancement are important for both effective human and machine use.


\begin{figure}[t]
\centering
\includegraphics[width=0.47\textwidth]{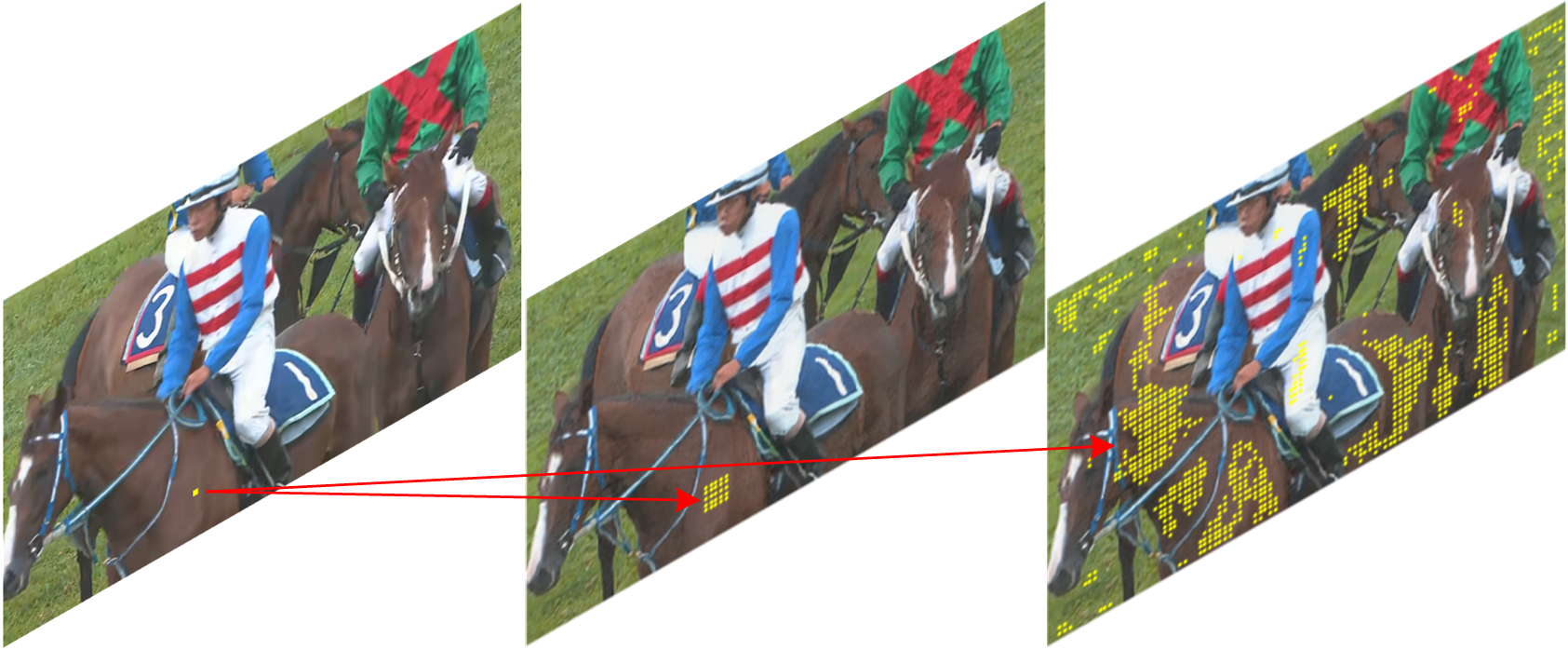}
\caption{Captured temporal information. (left) Target pixel in the current frame. (middle) Local temporal information captured by ``motion" in the adjacent frame. (right) Global temporal information captured by ``attention" in the adjacent frame.}
\label{fig 1}
\end{figure}

With the tremendous development of convolution neural networks (CNNs) in recent years, a large number of learning-based works have been proposed aiming at enhancing quality of compressed videos. Some of them \cite{b4,b5,b6,b7,b8,b9,b10} focus on enhancing the objective quality which is most commonly measured by peak signal to noise ratio (PSNR). Although the improvement of PSNR indicates the decrease of the objective distortion (e.g., mean square error (MSE)), it does not necessarily improve the visual experience \cite{b11}. Networks trained with adversarial loss, generative adversarial networks (GANs)\cite{b12}, are typically able to generate high-quality details and have significantly developed in the past few years, some algorithms \cite{b13,b14,b15,DCNGAN} focus on enhancing the perceptual quality of compressed videos with the help of GANs. However, they have not fully employed the global temporal information between consecutive frames to improve the perceptual quality of compressed videos.

Different from images, videos consist of consecutive frames. So in additional to spatial information, temporal information can be used for frame enhancement. How to make full use of global temporal information is a key issue in the perceptual quality enhancement of compressed videos. One of the most common methods to employ temporal information is motion compensation with estimated optical flows \cite{b8} \cite{b9}. However, estimated optical flows can hardly be accurate due to various artifacts in compressed frames, which penalizes the performance of temporal alignment. Another popular method to employ temporal information is to use deformable convolutions \cite{b16} \cite{b17} to achieve temporal alignment \cite{b10}. Compared with optical flows, deformable convolutions are more flexible and have potential to achieve better performance. However, both optical flows and deformable convolutions are based on motions between consecutive frames. In this case, only local temporal information can be employed due to the limited range of motion. As shown in Fig. 1 (middle), only local temporal information can be employed to enhance the pixel on given object (a horse, in this example). Although networks using optical flows and deformable convolutions to capture local temporal information are relatively fast, in specific use cases such as in video production or archiving, it is more important to get as high quality as possible, despite possible complexity increase. Since local areas will suffer from similar artifacts in video compression, it is of great importance to capture global information. Correlated pixels in the adjacent frame should be considered even though they are not in the reach of optical flow (which is coming from different tasks that do not look for such long range correlations), as shown in Fig. 1 (right). For this reason, the proposed network employs non-local attention modules to fully capture global temporal correlations in the consecutive frames and pave the way for the higher perceptual quality enhancement.

{Additional challenge for processing compressed videos is that at different quantization parameters (QPs) the videos have different characteristics. Intuitively, if one wants to process compressed videos at multiple QPs, multiple models will need to be correspondingly trained, stored and transmitted to adapt the characteristics of videos compressed at multiple QPs, which is extremely expensive in practical applications. To train only one model to adapt to multiple QPs with negligible performance loss, many QP-conditional adaptation mechanisms\cite{b25}\cite{b26}\cite{b28}\cite{b29} have been proposed for deep networks that address different problems. The strategy proposed in this paper is the first to successfully apply QP-conditional adaptation to perceptual quality enhancement of compressed videos, saving multiple sets of model parameters with almost the same enhancement performance compared to using the models trained for specific QPs.}

In this paper, we proposed an adaptation- and attention-based GAN, which is referred to as PeQuENet, to enhance the perceptual quality of compressed videos at various QPs by fully employing global temporal information in the consecutive frames. The proposed PeQuENet consists of four main parts: the pre-trained feature extraction module, the attention module, the progressive decoder module and the QP-conditional adaptation module.

Our main contributions are summarized as:
\begin{enumerate}
\item {With the goal to increase capability of learned solutions for perceptual quality enhancement of compressed videos, we integrate simplified non-local attention modules for conditional attention (i.e., fusing multi-dimensional features from two input sources - target and reference) with the aim of enhancing the target frame with the most relevant (or similar) parts of the reference frame and propose a set up where it outperforms previously used methods (i.e., optical flow and deformable convolution).}

\item For the first time, we successfully apply QP-conditional adaptation to compressed video perceptual quality enhancement network by feeding encoded QP information to the network to save model parameters without performance loss. 

\item We compare the proposed PeQuENet with state-of-the-art compressed video quality enhancement networks quantitatively and qualitatively. Experimental results demonstrated that the proposed PeQuENet consistently provides better quality frames.

\end{enumerate}

The remainder of the paper is organized as follows. Related work is summarized in Section II. The proposed PeQuENet for the perceptual quality enhancement of compressed videos is detailed in Section III. Experimental results are shown in Section IV to verify expected improvements when the proposed PeQuENet is deployed. Finally, Section V concludes the paper.

\section{Related Work}

\subsection{Quality Enhancement of Compressed Videos}
In the past few years, extensive works have been focused on the objective quality enhancement of compressed videos. Wang et al. \cite{b4} proposed a deep CNN with ten convolution layers and a residual structure to improve coding efficiency measured by BD-rate for HEVC. Dai et al. \cite{b5} employed a four-layer CNN with variable filter size and residual learning to improve performance of pose-processing in HEVC with low memory cost. Yang et al. \cite{b6} proposed models, referred to as QE-CNN-I and QE-CNN-P, to enhance the objective quality of I frames and P/B frames in compressed videos, respectively. To apply in time-constrained scenarios, they further proposed a scheme named TQEO to maximize the quality enhancement and meet the requirement of the complexity at the same time. Yang et al. \cite{b7} developed a detector based on Support Vector Machine (SVM) to recognize frames in peak quality and used them to improve neighboring frames in low quality. By employing local temporal information, they also alleviated quality fluctuation in compressed videos. Lu et al. \cite{b8} proposed a deep Kalman model to enhance quality of compressed videos. Instead of exploring temporal information in consecutive compressed frames, they used temporal information in consecutive restored frames in a recursive way, which ensured to reduce compression artifacts effectively. Guan et al. \cite{b9} further improved the algorithm in \cite{b7}. By using a Bidirectional Long Short-Term Memory (BiLSTM)-based detector, multi-scale feature extraction and densely connected mapping construction, they outperformed the methods in \cite{b4,b5,b6,b7,b8}. Deng et al. \cite{b10} proposed a fast quality enhancement network for compressed videos by incorporating deformable convolutions. With more flexibility, their proposed network achieves the state-of-the-art performance in terms of enhancing PSNR. Due to the high performance of the methods in \cite{b9} and \cite{b10}, they are used for comparison with the proposed network in experiments.

However, enhancing the objective quality of compressed videos does not necessarily improve visual experience of humans \cite{b11}. To improve QoE, many works have been proposed to enhance the perceptual quality of compressed videos. Wang et al. \cite{b13} proposed a multi-level wavelet-based GAN to enhance the perceptual quality of compressed videos. By recovering high-frequency sub-bands in the wavelet domain, their proposed network achieved enhanced perceptual quality. Wang et al. \cite{b14} proposed a simple perceptual quality enhancement network for HEVC compressed videos with the help of GAN and residual blocks. Jin et al. \cite{b15} proposed a multi-level progressive refinement network to enhance the perceptual quality for intra coding at the decoder end. By employing a coarse-to-fine refinement manner, their proposed network achieved trade-off between the quality and computational complexity. {Zhang et al. \cite{DCNGAN} proposed a deformable convolution-based GAN to improve the perception of compressed videos. After the current frame and its adjacent frames aligned with deformable convolutions, the perceptual quality of compressed videos was enhanced by a complex quality enhancement module.} \cite{b13,b14,b15,DCNGAN} are also used for comparison with the proposed network in experiments.

\begin{figure*}[t]
\centering
\includegraphics[width=0.97\textwidth]{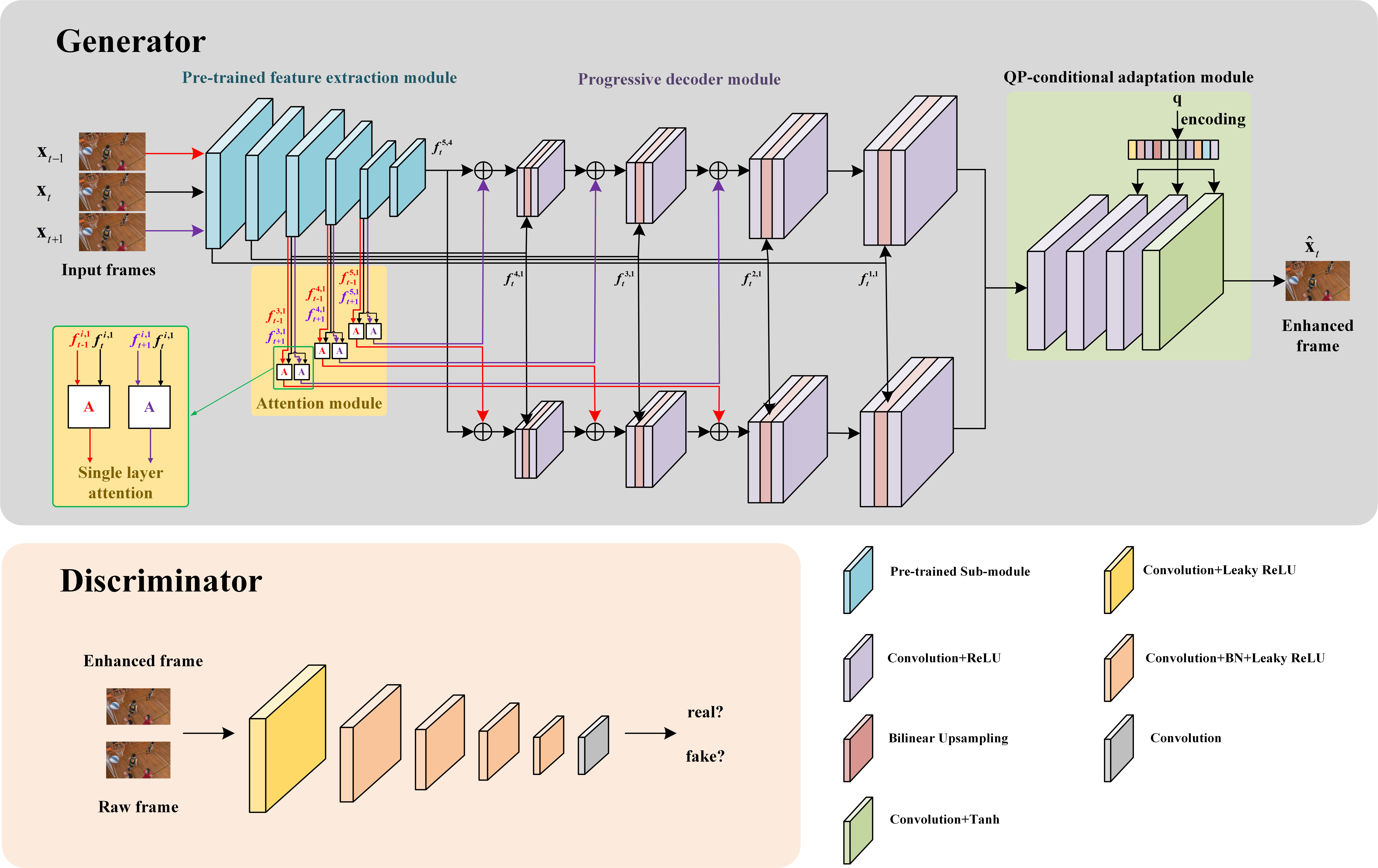}
\caption{{The proposed PeQuENet.}}
\label{fig 2}
\end{figure*}

\subsection{Non-Local Attention Mechanism}
Non-local attention mechanism \cite{b44} has been widely used in various applications to capture long-range correlations. Chen et al. \cite{b18} proposed an end-to-end deep image compression network based on non-local attention modules. By generating attention masks, they weighed features to achieve bit-rate allocation. Hu et al. \cite{b19} proposed a learning-based video compression network performing major operations in the feature space. With non-local attention blocks embedded in their proposed multi-frame feature fusion module, the coding efficiency has been greatly improved. Li et al. \cite{b20} proposed a pose-guided non-local attention-based GAN to achieve human pose transfer. Tan et al. \cite{b21} proposed a real-time Siamese tracking network. By exploring long-range dependencies between the target branch and the search branch, they captured important features in the target branch which were regarded as reliable guidance to the search branch. Wen et al. \cite{b22} proposed a medical image classification network using non-local attention mechanism to capture global information for better understanding the visual scene and identifying the tiny lesions. Li et al. \cite{b23} proposed a network for fashion landmark detection. With the help of the spatial-aware non-local attention mechanism, they utilized global spatial and semantic information to improve the detection performance. Blanch et al. \cite{b24} proposed an exemplar-based colorization network. By capturing global correlations between the target image and the reference image with attention modules at different resolutions, they achieved advanced style transfer. 

\subsection{QP-Conditional Adaption Mechanism}
Compressed videos encoded at different QPs have different characteristics, such as different degrees of artifacts. It is hard to achieve satisfactory performance for the network trained at one QP while employed at another. One way that seems straightforward is to train multiple models to adapt to various QPs. However, it inevitably requires large memory to store all model parameters at the decoder end. To save memory without performance loss, QP-conditional adaptation mechanism has been applied to deep networks addressing different problems. Liu et al. \cite{b25} proposed a QP-adaptive method and applied it to CNN-filters in video coding, improving coding efficiency with only 25\% of the parameters (typically the reduction in research papers is to 25\% because only four QPs are considered. But that with even more QPs which may be needed in practice, the number of models would be even bigger). Huang et al. \cite{b26} proposed a QP variable CNN-based in-loop filter for intra-coding of Versatile Video Coding (VVC) \cite{b27}. A QP attention module was designed and embedded into the residual block. With less model parameters, their proposed model achieved even better performance than that of QP-separate models. Song et al. \cite{b28} also proposed a CNN-based in-loop filter for intra-coding with QP-conditional adaptation. As one of inputs, normalized QP map was fed into the network to make it adaptive to various QPs. Zhang et al. \cite{b29} proposed a residual highway CNN for in-loop filtering in HEVC. By dividing the entire QP range into multiple QP bands, they trained one model for each band to save model parameters and maintain good performance.

\section{The Proposed PeQuENet}

The structure of the proposed PeQuENet is shown in Fig. 2. The proposed PeQuENet includes a generator (consisting of the pre-trained feature extraction module (fixed during the training), the attention module, the progressive decoder module and the QP-conditional adaptation module) and a discriminator. The generator and the discriminator are trained in an adversarial manner. The details of the architectures of the proposed PeQuENet is shown in our \textit{Supplemental Material}. The codes are available at \href{https://github.com/SaipingZhang/PeQuENet}{\textit{https://github.com/SaipingZhang/PeQuENet}}. 

To capture the temporal information in consecutive frames, we take the preceding frame $\mathbf{x}_{t - 1}$ and the succeeding frame $\mathbf{x}_{t + 1}$ as temporal reference frames to help enhance the perceptual quality of the target frame $\mathbf{x}_{t}$. Since we improve each compressed frame separately, a sliding-window strategy is employed to process the entire video. The proposed model can be expressed as

\begin{equation}
{\hat{\mathbf{x}}_t} = {f_\theta }\left( \mathbf{x}_{t - 1},\mathbf{x}_{t},\mathbf{x}_{t + 1} \right),
\end{equation}
where $\hat{\mathbf{x}}_t$ is the enhanced target frame (i.e., the output of the proposed PeQuENet). ${f_\theta }\left(   \cdot  \right)$ represents the proposed perceptual quality enhancement. $\theta$ is the set of the learnable model parameters. $\mathbf{x}_{t - 1}$, $\mathbf{x}_{t}$ and $\mathbf{x}_{t + 1}$ are three consecutive compressed frames (i.e., the inputs of the proposed PeQuENet). It should be noted that $\mathbf{x}_{t - 1} = \mathbf{x}_{t}$ when $\mathbf{x}_{t}$ is the first frame, and that $\mathbf{x}_{t + 1} = \mathbf{x}_{t}$ when $\mathbf{x}_{t}$ is the last frame in the video.

\subsection{Generator}

\subsubsection{Pre-Trained Feature Extraction Module}

Considering the strong ability of feature extraction of the VGG-19 network \cite{b30} pre-trained on ImageNet \cite{b31}, we employ it as the pre-trained feature extraction module in the proposed PeQuENet. 

Three consecutive compressed frames $\mathbf{x}_{t - 1}$, $\mathbf{x}_{t}$ and $\mathbf{x}_{t + 1}$ are separately fed into the pre-trained VGG-19 model to obtain features from multiple layers. Specifically, features of the target frame $\mathbf{x}_{t}$ are extracted from six layers of the pre-trained VGG-19 model, i.e., the first convolution layer before the $i$-th max-pooling (${l^{i,1}}$), for $i = 1, ..., 5$, and the fourth convolution layer before the fifth max-pooling (${l^{5,4}}$). Corresponding features are represented by $f_t^{i,1}$, $i=1,...,5$, and $f_t^{5,4}$. Features of the adjacent frame $\mathbf{x}_{j},j \in \left\{ {t - 1,t + 1} \right\}$ are extracted from the ${l^{3,1}}$, ${l^{4,1}}$ and ${l^{5,1}}$ layers, represented by $f_{j}^{3,1}$, $f_{j}^{4,1}$ and $f_{j}^{5,1}$.

As shown in Fig. 2, six feature pairs, i.e., $\left( {f_{j}^{i,1},f_t^{i,1}} \right)$, for each combination of $j \in \left\{ {t - 1,t + 1} \right\}$ and $i \in \left\{ {3,4,5} \right\}$ are fed into six attention blocks, to capture temporal correlations at different resolutions in the feature space. Furthermore, extracted features ${f_t^{i,1}}$, $i = 1, ....4$, and ${f_t^{5,4}}$ are transmitted to the proposed progressive decoder module to prevent the loss of information caused by downsampling at some extent.

\subsubsection{Attention Module}
As shown in Fig. 2, there are six attention blocks in the proposed attention module. Three of them receive feature pairs from the frames $\mathbf{x}_{t}$ and $\mathbf{x}_{t - 1}$ as inputs, and the other three receive feature pairs from the frames $\mathbf{x}_{t}$ and $\mathbf{x}_{t + 1}$ as inputs. Taking the attention block receiving $f_t^{i,1}$ and $f_{t - 1}^{i,1}$ as the inputs as an example, we show its structure in Fig. 3.

Specifically, $f_{t - 1}^{i,1} \in {\mathbb{R}^{c \times h \times w}}$ and $f_t^{i,1} \in {\mathbb{R}^{c \times h \times w}}$ ($h$ and $w$ are dimensions of feature maps at given level $i$) are fed into $1 \times 1$ convolution layers. Then they are reshaped to change the dimensions. After multiplication and \textit{softmax} operation, the correlation map in $hw \times hw$ is obtained to indicate the temporal correlation between $f_{t - 1}^{i,1}$ and $f_t^{i,1}$. Through multiplying the correlation map by the reshaped $f_{t - 1}^{i,1}$, the temporal information highly correlated to $f_{t}^{i,1}$ in $f_{t - 1}^{i,1}$ are amplified while the other information are suppressed. Finally, there is a reshaping operation to restore the dimension of the output, which prepares for adding the output to the corresponding features of the target frame $\mathbf{x}_{t}$ in the proposed progressive decoder module.

{Note that we integrate the simplified non-local attention mechanism for conditional attention (i.e., fusing multi-dimensional features from two input sources - target and reference). This differs from other papers (e.g.\cite{b44}) which use the same mechanism as a self-attention block. Specifically, the one in\cite{b44} substitutes convolution operations with self-attention blocks to capture long-range dependencies and hence, improve the performance of the original CNN for a particular task. However, our attention block in the proposed PeQuENet operates over pairs of consecutive frames and is used to compute analogies between target and reference frames with the aim of enhancing the target frame with the most relevant (or similar) parts of the reference frame. Besides, our attention block is simpler since features of just two frames (rather than a volume) are input as 2D spatial matrices (3D if considering channels) and used differently as key, value and query where the adjacent frame used as reference, and our attention block directly combines reference samples without the need of being projected, thus saving computational cost.}

\subsubsection{Progressive Decoder Module}

The proposed progressive decoder module takes extracted features of the target frame $\mathbf{x}_{t}$ and the outputs of attention blocks as the inputs. As shown in Fig. 2, this module has two branches. The first branch is designed for the frames $\mathbf{x}_{t}$ and $\mathbf{x}_{t+1}$, and the second branch is designed for the frames $\mathbf{x}_{t}$ and $\mathbf{x}_{t-1}$. Specifically, taking the first branch as an example, the output of the attention block is added to $f_t^{5,4}$. In this case, the temporal information highly correlated to the target frame $\mathbf{x}_{t}$ in the frame $\mathbf{x}_{t+1}$ is provided to the target frame $\mathbf{x}_{t}$ in the feature space. After that, the sum is fed into a $3 \times 3$ convolution layer. By Bilinear upsampling, the width and the height of feature maps are both doubled to cater for the resolution of $f_t^{4,1}$. After concatenating the upsampled feature maps and $f_t^{4,1}$ in the dimension of the channel, they are fed into another $3 \times 3$ convolution layer. As mentioned before,  $f_t^{4,1}$ can compensate lost information due to downsampling in the pre-trained feature extraction module at some extent. It should be noted that we employed Bilinear upsampling instead of the strided-deconvolution to increase the resolution of feature maps. This is because the strided-deconvolution has uneven overlap in the horizontal and vertical when the size of ``kernel" is not divisible by the ``stride", which sometimes results in severe checkerboard artifacts \cite{b32}. Similar operations are repeated until we obtain feature maps in the same resolution as that of the input frame. After two sets of feature maps from the first branch and the second branch are concatenated, they are fed into the QP-conditional adaptation module.

\subsubsection{QP-Conditional Adaptation Module}

The structure of the QP-conditional adaptation module is shown in Fig. 4. QP value $\mathbf{q}$ is fed into a one-hot encoding layer and a fully connected layer to be mapped into an encoded vector $ v\in {\mathbb{R}^{C \times 1 \times 1}}$. \textit{Softplus} is used as the activation function, as in \cite{b26}, since it ensures the positive outputs which are regarded as weights of features along the dimension of the channel. After performing element-wise multiplication between features $f\in {\mathbb{R}^{C \times H \times W}}$ and the encoded vector $v$, the weighed features are obtained and fed into the convolution layer. It should be noted that encoded $\mathbf{q}$ is repeatedly embedded into three convolution layer in the QP-conditional adaptation module to guarantee the proposed PeQuENet is successfully modulated by QP.

\subsection{Discriminator}

Considering the  perceptual quality enhancement of compressed videos mostly relies on the recovery of high-frequencies, it is sufficient for the discriminator to focus on local image patches to decide whether the generated high-frequencies are correct. In this case, the PatchGAN \cite{b33} is employed to assist in training the network to produce perceptually enhanced frames. Since the PatchGAN consists of convolution layers only, as shown in Fig. 2, its output is an array where each element represents the realness of the corresponding patch. After averaging the fidelity of all patches, the realness of the image is obtained. The advantage of the PatchGAN is that it processes each patch identically and independently, which indicates the PatchGAN has fewer parameters and lower computational complexity. 

Such approach is mathematically equivalent to cropping an image into multiple overlapping patches with appropriate sizes (i.e., the receptive fields of the PatchGAN) and employing a regular discriminator to process each of them independently.

\begin{figure}[t]
\centering
\includegraphics[width=0.48\textwidth]{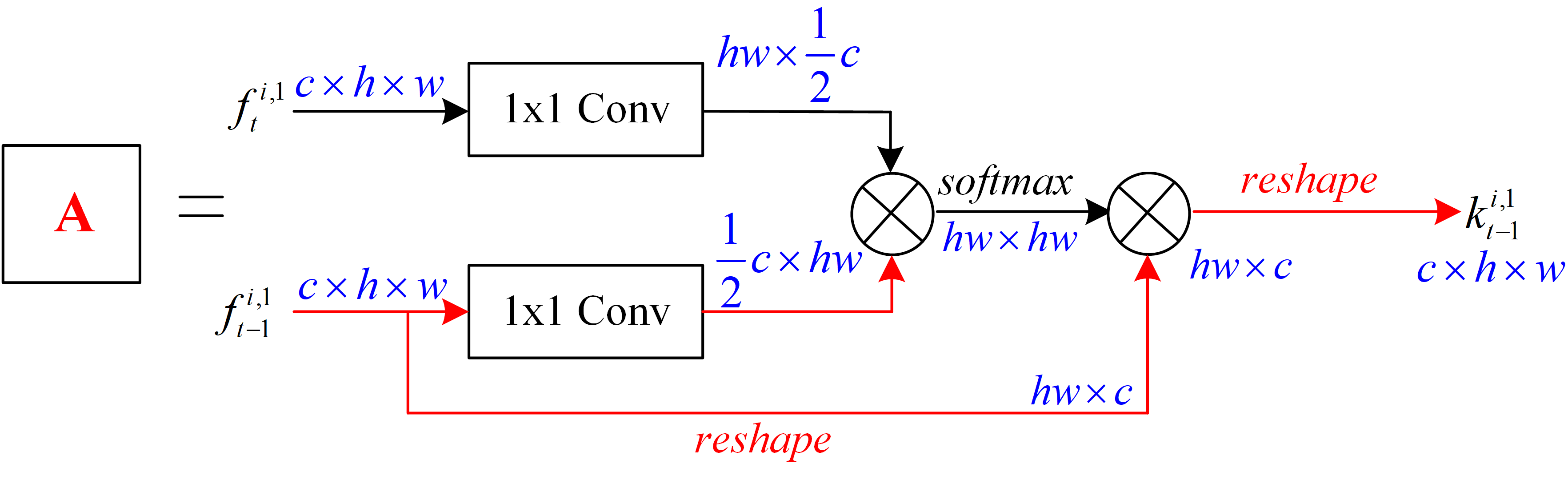}
\caption{The attention block.}
\label{fig 3}
\end{figure}

\begin{figure}[t]
\centering
\includegraphics[width=0.5\textwidth]{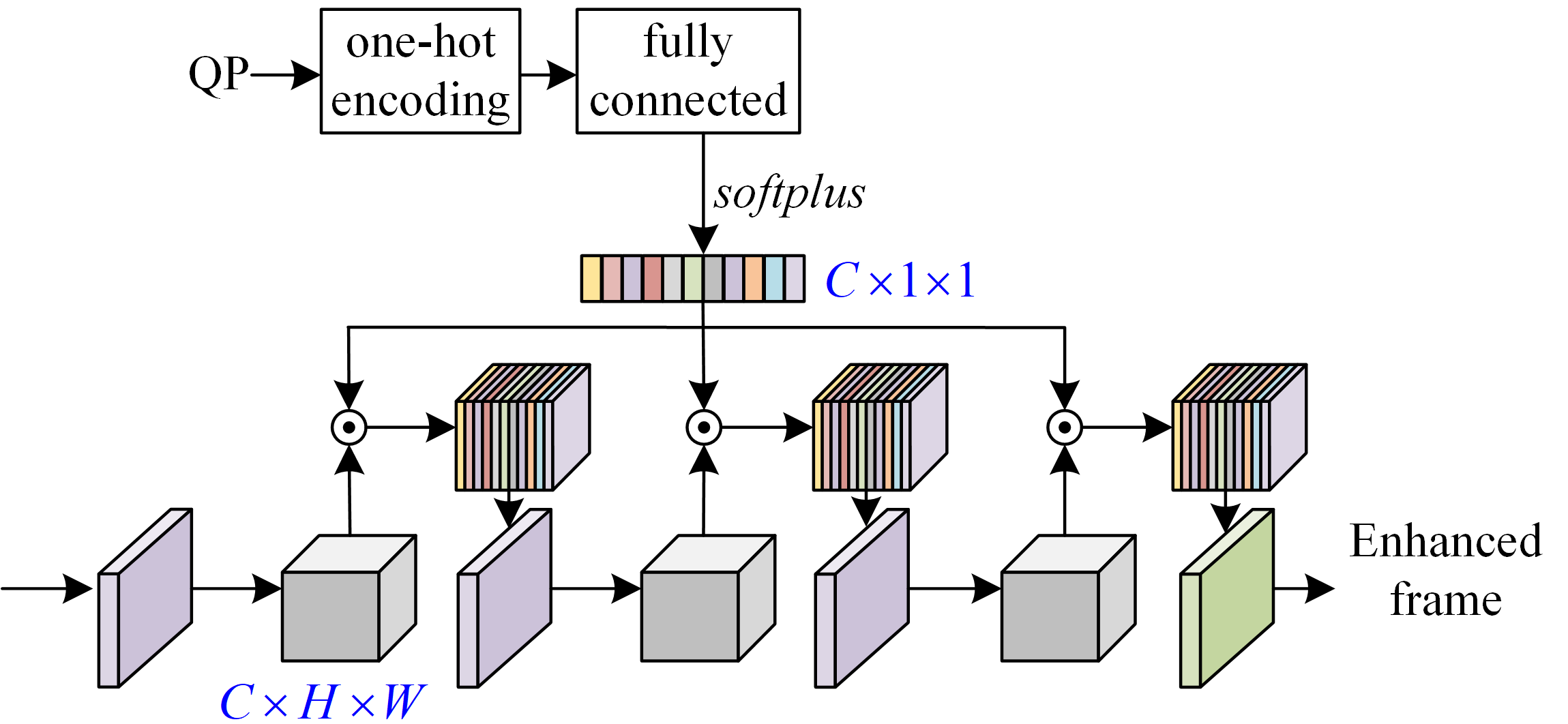}
\caption{The QP-conditional adaptation module.}
\label{fig 4}
\end{figure}

\subsection{Loss Functions}

The total loss $L_G$ of the generator of the proposed PeQuENet is the mixed loss which is formulated as the weighted sum of the adversarial loss ${L_{adv}}$, the VGG loss $L_{vgg}$ \cite{b34} and the feature matching loss $L_{fm}$ \cite{b35}: 

\begin{equation}
{L_G} = {{L_{adv}} + \alpha {L_{vgg}} + \beta {L_{fm}}},
\end{equation}
where $\alpha$ and $\beta$ are the corresponding weights. We employ $\alpha  = \beta  = 10$ in this paper.

The adversarial loss is referred to that proposed in LSGAN \cite{b36}:

\begin{equation}
{L_{adv}} = {\rm{\mathbb{E}}}_{\left({\mathbf{y},\mathbf{q}}\right)}\left[ {{{\left( {D\left( {G\left( \mathbf{y},\mathbf{q} \right)} \right) - 1} \right)}^2}} \right],
\end{equation}
where $\mathbf{y}= \left( \mathbf{x}_{t - 1},\mathbf{x}_{t},\mathbf{x}_{t + 1} \right)$ is the sequence of three consecutive input frames. $\mathbf{q}$ is the corresponding QP value. $G\left(  \cdot  \right)$ is the output of the generator. $D\left( \cdot \right)$ is the output of the discriminator.

The VGG loss is computed as
\begin{equation}
{{L_{vgg}}\left( G \right) = {\rm{\mathbb{E}}_{\left( {{\bf{y}},{\bf{q}},{\bf{x}}} \right)}}\sum\limits_{i{\rm{ = 1}}}^{{N_f}} {\frac{{\rm{1}}}{{{M_f}}}\sum\limits_{j{\rm{ = 1}}}^{{M_f}} {{{\left\| {f_j^i\left( {\bf{x}} \right){\rm{ - }}f_j^i\left( {G\left( {{\bf{y}},{\bf{q}}} \right)} \right)} \right\|}_{\rm{1}}}}},}
\end{equation}
where ${f^i_j}\left(\cdot  \right)$ represents the $j$-th spatial element of the output tensor of the $i$-th layer from a pre-trained VGG-19 model. {$M_f$} and $N_f$ are the number of spatial elements and the number of layers, respectively. $\mathbf{x}$ is the corresponding raw frame, i.e., the original frame before compression.

Similarly, the feature matching loss is computed as

\setlength\tabcolsep{1.5pt}
\begin{table*}[htbp]
\caption{{Overall Performance on LPIPS and DISTS of JCT-VC Standard Test Sequences of H.265/HEVC}}
\begin{center}
\begin{threeparttable}[b]
\begin{tabular}{ccccccccccccccccccc}
\toprule
{\multirow{2}{*}{QP}}                   & \multicolumn{2}{c}{\multirow{2}{*}{Sequences\tnote{*}}} & \multicolumn{2}{c}{Compressed}  & \multicolumn{2}{c}{MFQE 2.0 \cite{b9}} & \multicolumn{2}{c}{STDF \cite{b10}} & \multicolumn{2}{c}{MW-GAN \cite{b13}} & \multicolumn{2}{c}{VPE-GAN \cite{b14}} & \multicolumn{2}{c}{MPRNet \cite{b15}} &
\multicolumn{2}{c}{{DCNGAN \cite{DCNGAN}}} &\multicolumn{2}{c}{Proposed} \\
\multirow{20}{*}{32} & \multicolumn{2}{c}{}                           & LPIPS         & DISTS    & LPIPS         & DISTS    & LPIPS       & DISTS      & LPIPS        & DISTS       & LPIPS        & DISTS        & LPIPS        & DISTS       & {LPIPS}         & {DISTS}   & LPIPS         & DISTS     \\
\midrule
                     & \textit{Class A}                    & \textit{Traffic}   &0.170       &0.014       &0.184               &0.014              &0.094             &0.009            &0.138              &--            &0.179              &0.029               &0.074              &0.016               & {0.070}      & {\textbf{0.006}}      &\textbf{0.058}               &\textbf{0.006}              \\
                     &                            & \textit{PeopleOnStreet}     &0.150       &0.018       &0.167               &0.018              &0.133             &0.010            &0.130              &--             &0.135              &0.015              &0.096              &0.013                  &{0.086}       & {0.008}    &\textbf{0.074}               &\textbf{0.007}              \\
                     & \multirow{5}{*}{\textit{Class B}}   & \textit{Kimono}    &0.258       &0.043       &0.294               &0.046              &0.160             &0.026            &0.189              &--             &0.180              &0.034              &0.136              &0.053            &  {0.108}     &{0.023}           &\textbf{0.107}               &\textbf{0.021}              \\
                     &                            & \textit{ParkScene}          &0.276       &0.044       &0.286               &0.045              &0.182             &0.027            &0.244              &--             &0.196              &0.037              &0.140              &0.039            &  {0.123}     &{0.023}           &\textbf{0.101}               &\textbf{0.021}              \\
                     &                            & \textit{Cactus}             &0.260       &0.022       &0.288               &0.022              &0.136             &0.012            &0.151              &--             &0.126              &0.017              &0.110              &0.020              &   {0.096}    & {\textbf{0.010}}       &\textbf{0.087}               &\textbf{0.010}              \\
                     &                            & \textit{BQTerrace}          &0.215       &0.032       &0.241               &0.034              &0.152             &0.021            &0.116              &--             &0.140              &0.040              &0.129              &0.042              & {0.113}      & {0.018}        &\textbf{0.090}               &\textbf{0.013}              \\
                     &                            & \textit{BasketballDrive}    &0.247       &0.028       &0.279               &0.031              &0.166             &0.022            &0.141              &--             &0.132              &0.025              &0.137              &0.027             & {0.099}      &{0.015}          &\textbf{0.092}               &\textbf{0.014}              \\
                     & \multirow{4}{*}{\textit{Class C}}   & \textit{RaceHorses}&0.147       &0.066       &0.174               &0.075              &0.120             &0.061            &0.126              &--             &0.101              &0.055              &0.083              &0.051               &{0.089}       & {0.042}       &\textbf{0.069}               &\textbf{0.039}              \\
                     &                            & \textit{BQMall}             &0.124       &0.066       &0.145               &0.071              &0.089             &0.050            &0.091              &--             &0.112              &0.063              &0.076              &0.056              & {0.072}      &  {0.038}       &\textbf{0.051}               &\textbf{0.036}              \\
                     &                            & \textit{PartyScene}         &0.101       &0.057       &0.126               &0.060              &0.067             &0.042            &\textbf{0.026}     &--             &0.091              &0.045              &0.056              &0.038               & {0.075}      & {0.029}       &0.040                      &\textbf{0.024}              \\
                     &                            & \textit{BasketballDrill}    &0.156       &0.073       &0.181               &0.079              &0.126             &0.068            &0.109              &--             &0.105              &0.060              &0.085              &0.054                 & {0.072}      &  {0.040}    &\textbf{0.061}               &\textbf{0.037}              \\
                     & \multirow{4}{*}{\textit{Class D}}&\textit{\textit{RaceHorses}}&0.122  &0.121       &0.143               &0.132              &0.098             &0.113            &0.117              &--             &0.093              &0.126              &0.069              &0.097            &   {0.072}    &{0.091}          &\textbf{0.052}                &\textbf{0.078}              \\
                     &                            & \textit{BQSquare}           &0.110       &0.150       &0.121               &0.160              &0.084             &0.130            &0.073              &--             &0.066              &0.112              &0.068              &0.122            &  {0.104}     &{0.123}           &\textbf{0.047}               &\textbf{0.084}              \\
                     &                            & \textit{BlowingBubbles}     &0.102       &0.117       &0.111               &0.128              &0.068             &0.104            &0.063              &--             &0.072              &0.096              &0.054              &0.078              & {0.065}      &  {0.084}       &\textbf{0.037}               &\textbf{0.060}              \\
                     &                            & \textit{BasketballPass}     &0.116       &0.135       &0.135               &0.150              &0.099             &0.127            &0.095              &--             &0.085              &0.116              &0.074              &0.107               & {0.067}     &  {0.099}      &\textbf{0.056}               &\textbf{0.086}              \\
                     & \multirow{3}{*}{\textit{Class E}}   & \textit{FourPeople} &0.120      &0.037       &0.128               &0.038              &0.089             &0.022            &0.080              &--             &0.103              &0.028              &0.076              &0.026            & {0.054}      &{0.016}           &\textbf{0.051}               &\textbf{0.013}              \\
                     &                            & \textit{Johnny}             &0.148       &0.035       &0.159               &0.035              &0.111             &0.021            &0.083              &--             &0.178              &0.059              &0.186              &0.060           & {0.063}      & {0.014}           &\textbf{0.050}               &\textbf{0.013}              \\
                     &                            & \textit{KristenAndSara}    &0.134        &0.038       &0.148               &0.039              &0.106             &0.025            &0.108              &--             &0.136              &0.046              &0.140              &0.059           &  {0.062}     &{0.019}            &\textbf{0.050}               &\textbf{0.017}              \\
                     & \multicolumn{2}{c}{\textit{Average}}                    &0.164        &0.061       &0.184               &0.065              &0.116             &0.049            &0.115              &--             &0.124              &0.056              &0.099              &0.053           &{0.083}       &{0.039}            &\textbf{0.065}               &\textbf{0.032}              \\
\hline
22                   & \multicolumn{2}{c}{\textit{Average}}                    &0.077        &0.020  &0.087                    &0.022                   &0.050        &0.014       &--                   &--             &0.097              &0.047              &0.037              &0.014                  & {0.042}      & {0.017}    &\textbf{0.026}               &\textbf{0.009}              \\

27                   & \multicolumn{2}{c}{\textit{Average}}                    &0.116        &0.037  &0.130                    &0.040                   &0.077        &0.029       &--                   &--             &0.103              &0.054              &0.057              &0.026               &  {0.059}     &  {0.026}      &\textbf{0.040}               &\textbf{0.018}              \\
    
37                   & \multicolumn{2}{c}{\textit{Average}}                    &0.223        &0.089  &0.232               &0.086              &0.168        &0.080       &0.177              &--             &0.148              &0.070              &0.147              &0.070             & {0.120}      &  {0.058}        &\textbf{0.106}               &\textbf{0.057}              \\
\bottomrule 
\end{tabular}
\begin{tablenotes}
\item[*] Video Resolution: Class A (2560x1600), Class B (1920x1080), Class C (832x480), Class D (416x240), Class E (1280x720)
\end{tablenotes}
\end{threeparttable}
\end{center}
\end{table*}

\begin{equation}
{{L_{fm}}\left( {G,D} \right) = {\rm{\mathbb{E}}_{\left( {{\bf{y}},{\bf{q}},{\bf{x}}} \right)}}\sum\limits_{i{\rm{ = 1}}}^{{N_g}} {\frac{{\rm{1}}}{{M_g}}\sum\limits_{j{\rm{ = 1}}}^{{M_g}} {{{\left\| {g_j^i\left( {\bf{x}} \right){\rm{ - }}g_j^i\left( {G\left( {{\bf{y}},{\bf{q}}} \right)} \right)} \right\|}_{\rm{1}}}} },}
\end{equation}
where ${g^i_j}\left(\cdot  \right)$ represents the $j$-th spatial element of the output tensor of the $i$-th layer selected from the discriminator. {$M_g$} and $N_g$ are the number of spatial elements and the number of layers, respectively.

The loss of the discriminator of the proposed PeQuENet is also referred to that proposed in LSGAN:

\begin{equation}
L_D = \frac{1}{2}{\rm{\mathbb{E}}}_{\left({\mathbf{y},\mathbf{q}}\right)}\left[ {{{\left( {D\left( {G\left( \mathbf{y},\mathbf{q} \right)} \right)} \right)}^2}} \right] + \frac{1}{2}{\rm{\mathbb{E}}}_{\mathbf{x}}\left[ {D{{\left(\left( \mathbf{x} \right)-1\right)}^2}} \right].
\end{equation}

The generator and the discriminator are trained in an adversarial manner, which indicates that $L_G$ and $L_D$ are minimized alternatively.

\section{Experimental Results}
\subsection{Datasets}
 
108 sequences representing various content types and resolutions introduced in \cite{b9} are employed for training. All sequences are compressed by H.265/HEVC reference software HM16.5 under Low Delay P (LDP) configuration with QPs set to be 22, 27, 32 and 37. It should be noted that our training dataset includes compressed sequences at all of the four QPs to give the proposed PeQuENet the ability of QP-conditional adaptation. Raw sequences and compressed sequences are cropped into non-overlapping clips in $128 \times 128$ as training samples. Before fed into the network, all training samples are shuffled.

For evaluating the performance, tests are performed on Joint Collaborative Team on Video Coding (JCT-VC) \cite{b37} standard test sequences. All test sequences are also compressed by HM 16.5 under LDP configuration at the four QPs.

\subsection{Implementation Details}
Adam optimizer \cite{b38} with ${\beta _1} = 0.9$, ${\beta _2} = 0.999$ and ${\epsilon} = {10^{ - 8}}$ is adopted to train the model. Learning rate is set to be ${10^{ - 4}}$ thorough out the training process. Batch size is 32. For fair comparison with previous work, we only train our model on luminance component, but the proposed algorithm can be also extended to chrominance component. Perceptual quality metrics, i.e., LPIPS \cite{b39} and DISTS \cite{b40}, are used to evaluate the performance of the proposed PeQuENet. Smaller LPIPS/DISTS values correspond to better performance. It should be noted that we only train one model to adapt to videos compressed at multiple QPs.

\begin{figure*}[h]
\centering
\subfloat[\textit{{Johnny}}]{\includegraphics[width=0.47\textwidth]{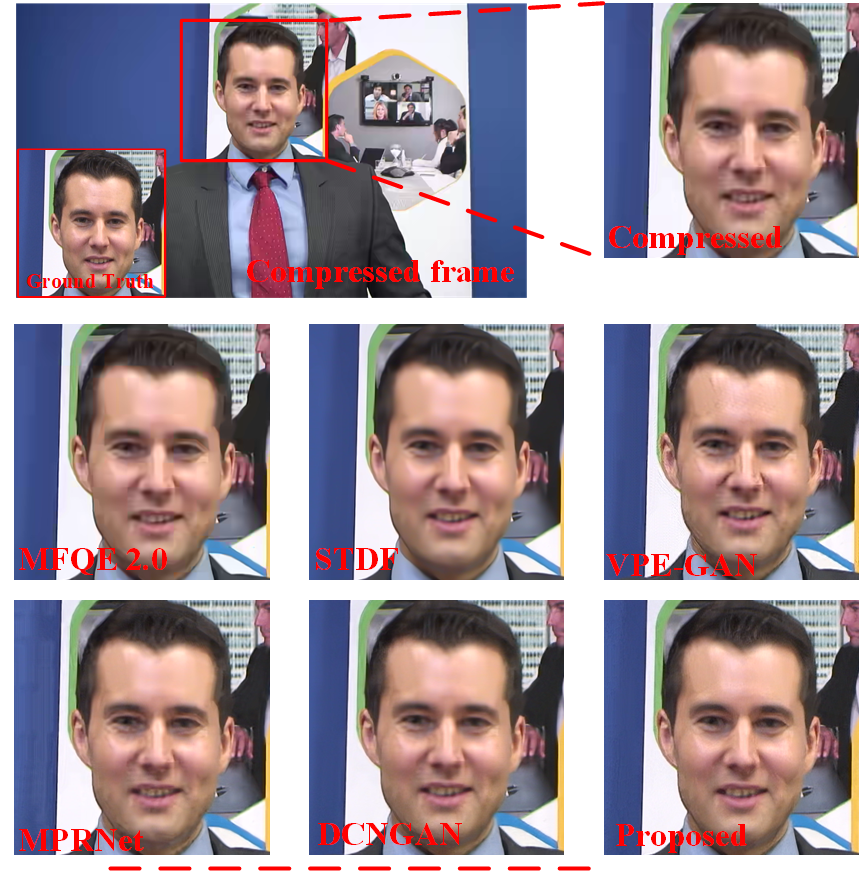}
\label{fig 5 (a)}}
\hfil
\subfloat[\textit{{PartyScene}}]{\includegraphics[width=0.46\textwidth]{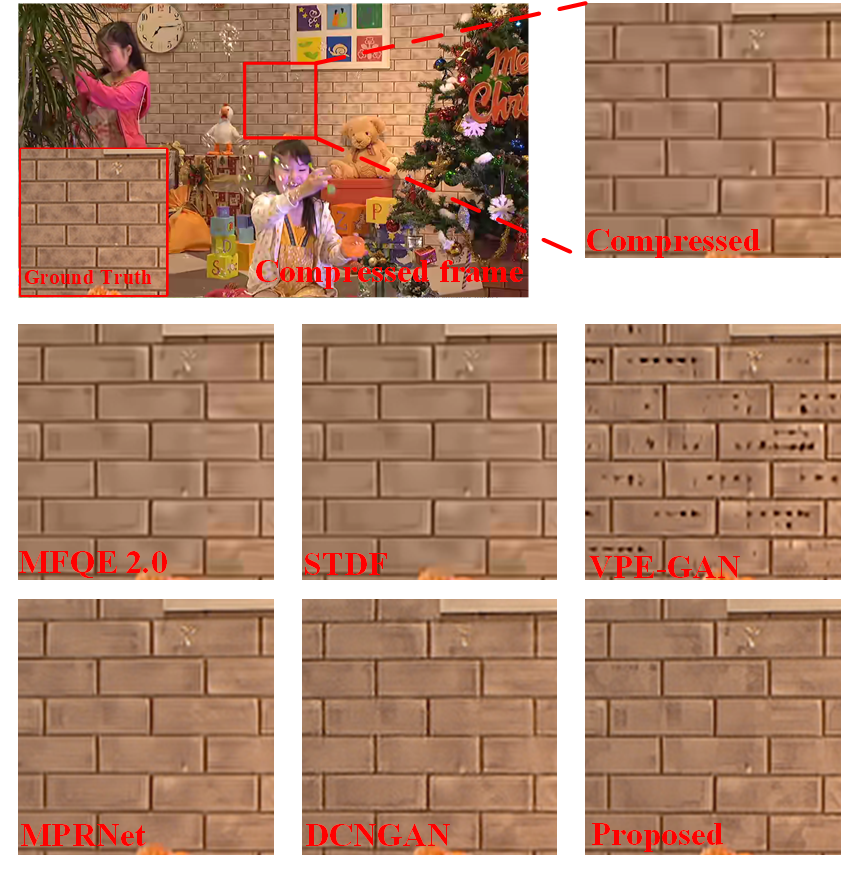}
\label{fig 5 (b)}}
\hfil
\subfloat[\textit{{RaceHorses}}]{\includegraphics[width=0.47\textwidth]{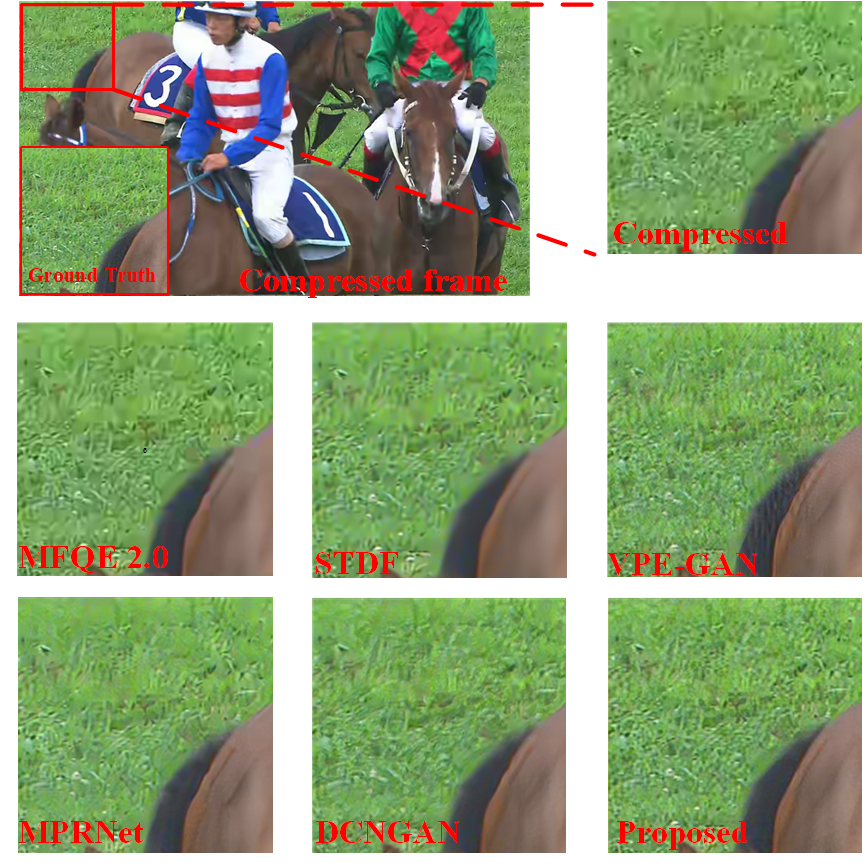}
\label{fig 5 (b)}}
\hfil
\subfloat[\textit{{ParkScene}}]{\includegraphics[width=0.46\textwidth]{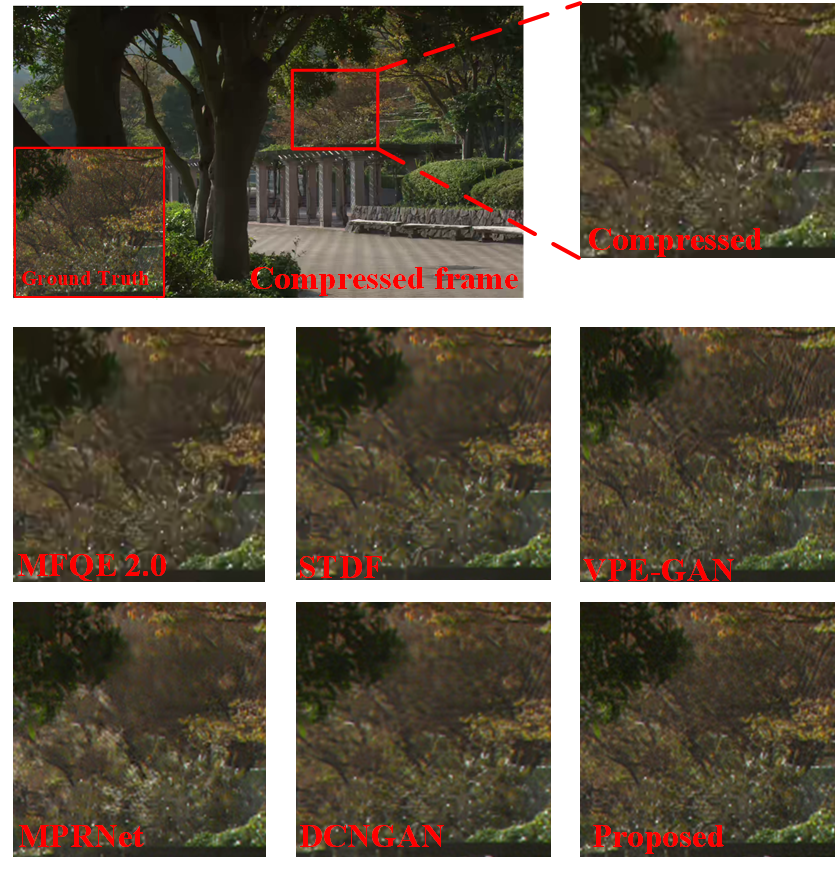}
\label{fig 5 (b)}}
\caption{{Qualitative comparison of the proposed PeQuENet with the other video enhancement algorithms.}}
\label{fig 5}
\end{figure*}

\setlength\tabcolsep{4pt}
\begin{table*}[htbp]
\caption{LPIPS Performance of Four Models Independently Trained for Four Different QPs}
\begin{center}
\begin{tabular}{lccccccccccccccccc}
\toprule
\multicolumn{2}{c}{\multirow{2}{*}{Sequences}}      & \multicolumn{4}{c}{Trained\_QP22}            & \multicolumn{4}{c}{Trained\_QP27}              & \multicolumn{4}{c}{Trained\_QP32}            & \multicolumn{4}{c}{Trained\_QP37}               \\
\multicolumn{2}{l}{}                                &22 &27 &32 &37 &22 &27    &32 &37 &22 &27 &32 &37 &22 &27 &32 &37    \\
\midrule
\multirow{2}{*}{\textit{Class A}} & \textit{Traffic}&0.018        &0.036             &0.069             &0.117             &0.024        &0.032           &0.059          &0.108          &0.046        &0.043        &0.057       &0.097        &0.087        &0.076        &0.073        &0.095          \\
                         & \textit{PeopleOnStreet}  &0.021        &0.044             &0.094             &0.148             &0.030        &0.037           &0.082          &0.143          &0.052        &0.049        &0.080       &0.137        &0.090        &0.079        &0.088        &0.127          \\
\multirow{5}{*}{\textit{Class B}} & \textit{Kimono} &0.069        &0.088             &0.117             &0.159             &0.063        &0.074           &0.103          &0.150          &0.087        &0.088        &0.111       &0.148        &0.115        &0.110        &0.116        &0.145          \\
                         & \textit{ParkScene}       &0.039        &0.076             &0.134             &0.205             &0.037        &0.054           &0.103          &0.179          &0.083        &0.075        &0.098       &0.162        &0.163        &0.133        &0.121        &0.155          \\
                         & \textit{Cactus}          &0.043        &0.075             &0.103             &0.143             &0.042        &0.058           &0.084          &0.126          &0.065        &0.068        &0.085       &0.124        &0.109        &0.099        &0.099        &0.122          \\
                         & \textit{BQTerrace}       &0.033        &0.078             &0.109             &0.145             &0.030        &0.063           &0.093          &0.131          &0.052        &0.075        &0.095       &0.129        &0.087        &0.089        &0.097        &0.122          \\
                         & \textit{BasketballDrive} &0.048        &0.078             &0.109             &0.146             &0.045        &0.063           &0.093          &0.131          &0.072        &0.075        &0.094       &0.127        &0.087        &0.084        &0.096        &0.123          \\
\multirow{4}{*}{\textit{Class C}} & \textit{RaceHorses}&0.026     &0.048             &0.081             &0.143             &0.025        &0.040           &0.070          &0.131          &0.043        &0.047        &0.069       &0.112        &0.099        &0.091        &0.095        &0.116          \\
                         & \textit{BQMall}          &0.015        &0.033             &0.062             &0.111             &0.015        &0.027           &0.054          &0.102          &0.026        &0.032        &0.051       &0.095        &0.049        &0.049        &0.060        &0.093          \\
                         & \textit{PartyScene}      &0.006        &0.023             &0.054             &0.117             &0.008        &0.017           &0.045          &0.105          &0.015        &0.019        &0.041       &0.095        &0.050        &0.037        &0.047        &0.087          \\
                         & \textit{BasketballDrill} &0.014        &0.037             &0.087             &0.134             &0.017        &0.029           &0.068          &0.122          &0.043        &0.039        &0.062       &0.106        &0.080        &0.067        &0.072        &0.113          \\
\multirow{4}{*}{\textit{Class D}} & \textit{RaceHorses}&0.009     &0.024             &0.068             &0.144             &0.014        &0.022           &0.056          &0.131          &0.033        &0.031        &0.050       &0.116        &0.091        &0.080        &0.073        &0.103          \\
                         & \textit{BQSquare}        &0.006        &0.025             &0.064             &0.103             &0.008        &0.020           &0.056          &0.093          &0.017        &0.024        &0.048       &0.085        &0.042        &0.043        &0.057        &0.085          \\
                         & \textit{BlowingBubbles}  &0.006        &0.018             &0.054             &0.128             &0.007        &0.014           &0.043          &0.111          &0.015        &0.018        &0.036       &0.096        &0.046        &0.041        &0.046        &0.087          \\
                         & \textit{BasketballPass}  &0.012        &0.033             &0.071             &0.136             &0.013        &0.026           &0.061          &0.125          &0.029        &0.032        &0.055       &0.112        &0.057        &0.053        &0.065        &0.110          \\
\multirow{3}{*}{\textit{Class E}} & \textit{FourPeople}&0.028     &0.040             &0.057             &0.087             &0.026        &0.032           &0.047          &0.078          &0.042        &0.045        &0.054       &0.082        &0.054        &0.053        &0.058        &0.076          \\
                         & \textit{Johnny}          &0.031        &0.044             &0.059             &0.085             &0.030        &0.037           &0.050          &0.077          &0.034        &0.039        &0.050       &0.079        &0.050        &0.055        &0.062        &0.078          \\
                         & \textit{KristenAndSara}  &0.040        &0.056             &0.069             &0.096             &0.030        &0.038           &0.055          &0.084          &0.044        &0.045        &0.058       &0.082        &0.056        &0.055        &0.058        &0.075          \\
\multicolumn{2}{c}{Average}                         &\textbf{0.026}        &0.048             &0.081             &0.130             &0.026        &\textbf{0.038}           &0.068          &0.118          &0.044        &0.047        &\textbf{0.066}       &0.110        &0.078        &0.072        &0.077        &\textbf{0.106}          \\
\bottomrule
\end{tabular}
\end{center}
\end{table*}

\subsection{Evaluation of Video Quality Enhancement}
We compare the proposed PeQuENet with the state-of-the-art compressed video quality enhancement networks, i.e., the MFQE 2.0 \cite{b9}, STDF \cite{b10} \cite{b43}, MW-GAN \cite{b13}, VPE-GAN \cite{b14}, MPRNet \cite{b15} {and DCNGAN \cite{DCNGAN}}. Among them, the MFQE 2.0 and STDF were proposed to enhance the objective quality (i.e., PSNR) of compressed videos while the MW-GAN, VPE-GAN, MPRNet {and DCNGAN} were trained with the help of GANs to enhance the perceptual quality of compressed videos. For fair comparison, all networks are trained on the same dataset (including the MW-GAN $\footnote{Since our retrained MW-GAN has not achieved the performance presented in [13] and their pre-trained models for the four QPs have not been published, the performance of the MW-GAN shown in TABLE I is directly taken from [13] for fair comparison. It should be noted that the performance shown in [13] is measured on $\Delta$LPIPS which is the difference between the LPIPS of the output and the LPIPS of the input. For better illustration, we simply further calculate LPIPS of the output of MW-GAN by adding the $\Delta$LPIPS to the LPIPS of the input (the input of the MW-GAN is the same as that of the proposed PeQuENet) and show the results in TABLE I.}$ detailed in \cite{b13}). The only difference is that {the DCNGAN} and the proposed PeQuENet are trained only once with samples compressed at the four QPs, while the other networks use samples compressed at a single QP to train individual models. It should be noted that, when compared with the STDF, we choose their proposed model taking seven consecutive frames as inputs since it achieved better performance than that of their proposed model fed with three consecutive frames by leveraging more temporal information. The overall quantitative performance measured by LPIPS and DISTS is shown in TABLE I where the performance of videos compressed by HM 16.5 (i.e., the input of the proposed network) is shown in the first two columns for a reference.

Specifically, the MFQE 2.0, which was designed for the objective quality enhancement, fails to improve the perceptual quality of compressed videos measured with both LPIPS and DISTS. The STDF achieves perceptual quality enhancement although it was originally also proposed for the objective quality enhancement. Note that this discrepancy between subjective and objective results for the MFQE 2.0 and the STDF indicates the inconsistency between subjective quality and objective quality metrics. On the other hand, although the VPE-GAN can enhance the perceptual quality at higher QPs (i.e., QP 32 and QP 37) to some extent, it fails at lower QPs (i.e., QP 22). As for the MW-GAN, MPRNet {and DCNGAN}, they can enhance the perceptual quality at all tested QPs with the help of GANs, but their performance is unsatisfactory in terms of slight decrease of LPIPS and DISTS. With QP-conditional adaptation, one trained model of the proposed PeQuENet is tested on the four QPs achieving the best performance and saving 75\% model parameters at the same time.

Taking four frames in test sequences compressed at QP 37 as examples, we show qualitative comparison of video enhancement algorithms in Fig. 5. Overall, the MFQE 2.0 and the STDF produce overly smooth results. While smoothness can be desirable in the background, as shown in Fig. 5 (b), but it penalizes the perceptual quality of the areas with complex texture, as shown in Fig. 5 (a), (c) and (d). The VPE-GAN tends to generate severe artifacts in the areas with simple texture to combat with over-smoothness, as shown in Fig. 5 (b), which leads to visual experience deterioration. The MPRNet and {the DCNGAN} can enhance perceptual quality to some extent, but the problem of over-smoothness has not been solved completely, as shown in Fig. 5 (c) and (d). Compared with the other video enhancement algorithms, the proposed PeQuENet not only generates more realistic details in the areas with complex texture to eliminate over-smoothness, but also avoids the appearance of artifacts in the areas with simple texture, which illustrates the best performance of the proposed PeQuENet.

{To further evaluate the subjective quality of the proposed method, a mean opinion score (MOS) test is also conducted at QP 37. Specifically, 15 subjects have rated scores from 1 to 5 following\cite{b45} on 18 JCT-VC standard test sequences of H.265/HEVC enhanced by the VPE-GAN, MPRNet, DCNGAN and the proposed PeQuENet. Higher scores correspond to better subjective quality. Average MOS results of all methods are compared in Fig. 6. As can be seen, the proposed PeQuENet achieves the highest average MOS.}

\begin{figure}[]
\centering
\includegraphics[width=0.4\textwidth]{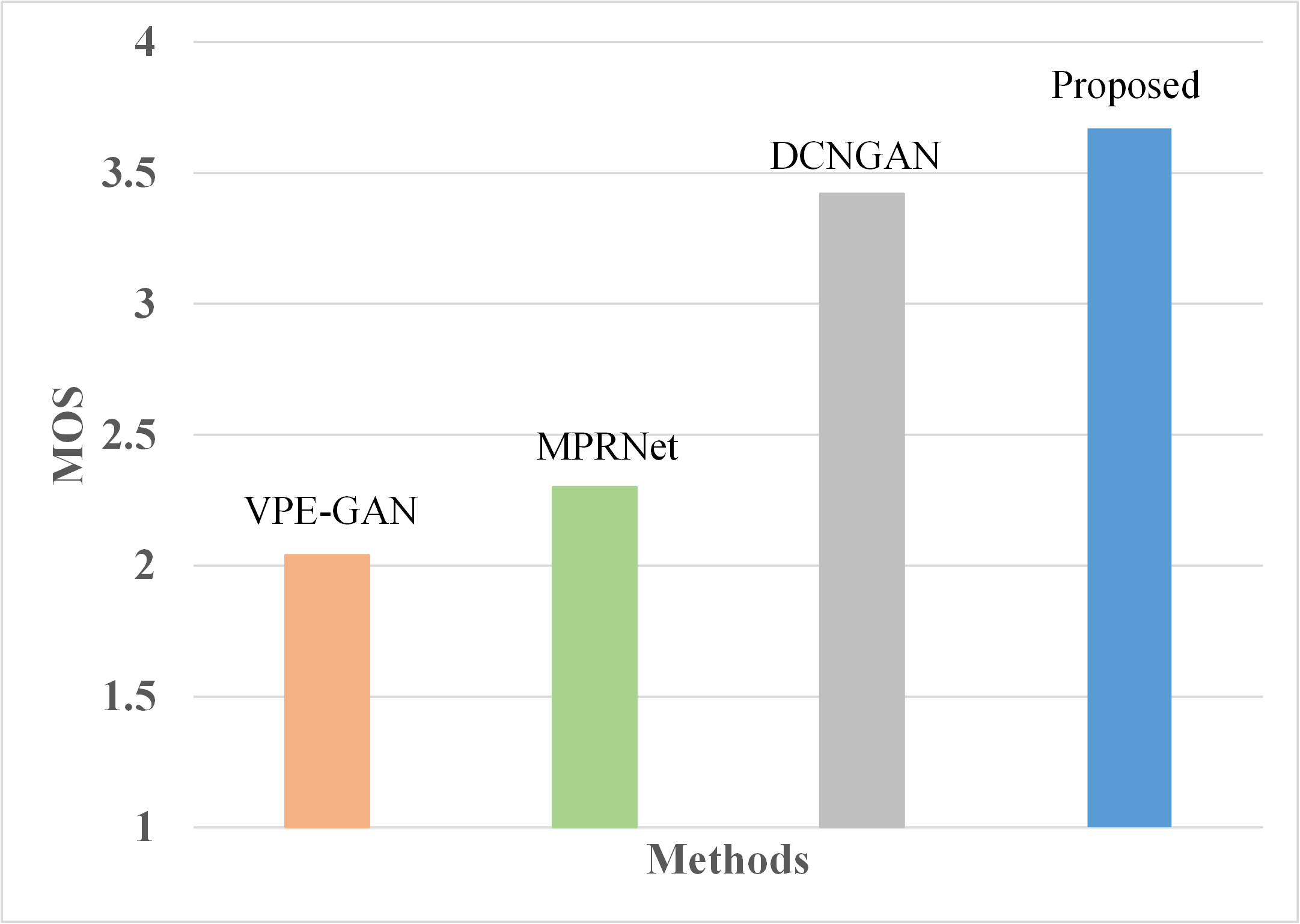}
\caption{{Average MOS results comparison.}}
\label{fig 6}
\end{figure}

{GANs help improve the QoE by training the network to generate high-frequency information. Hence, when GANs are employed in the video perceptual quality enhancement networks, the temporal consistency should be examined in case that generated details in each frame are inconsistent. Compared with three perceptual quality enhancement algorithms utilizing GANs, i.e., the VPE-GAN, MPRNet and DCNGAN, temporal consistency performance of the sequence \textit{BQSquare} compressed at QP 37 is shown in Fig. 7 as an example. There is significant noise in the VPE-GAN and the DCNGAN, which indicates flicking artifacts in the enhanced videos. As for the MPRNet, it performs better but still contains temporal inconsistency. By fully employing global temporal information in the consecutive frames, the proposed PeQuENet shows the smoothest temporal transition in the output videos.}

\begin{figure}[]
\centering
\includegraphics[width=0.48\textwidth]{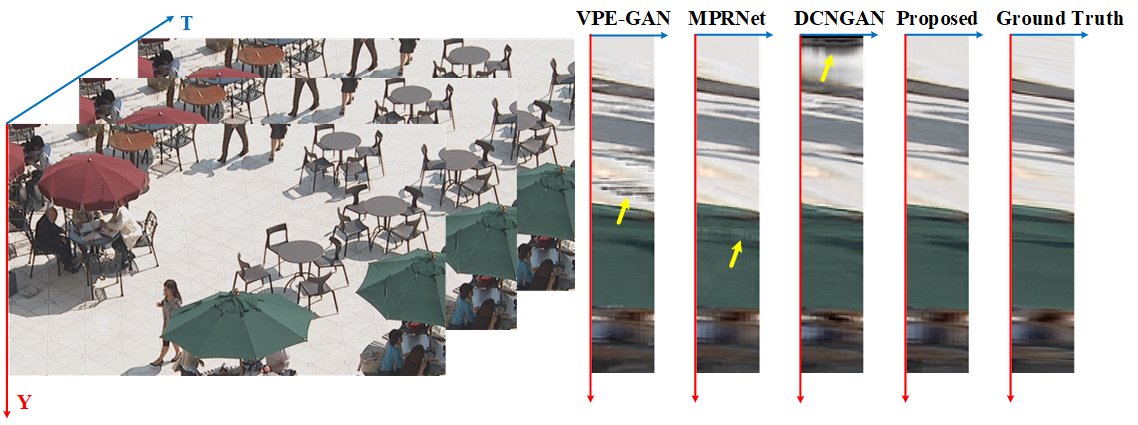}
\caption{{Temporal consistency comparison of the proposed PeQuENet with the VPE-GAN, MPRNet and DCNGAN.}}
\label{fig 7}
\end{figure}

\begin{figure}[]
\centering
\includegraphics[width=0.5\textwidth]{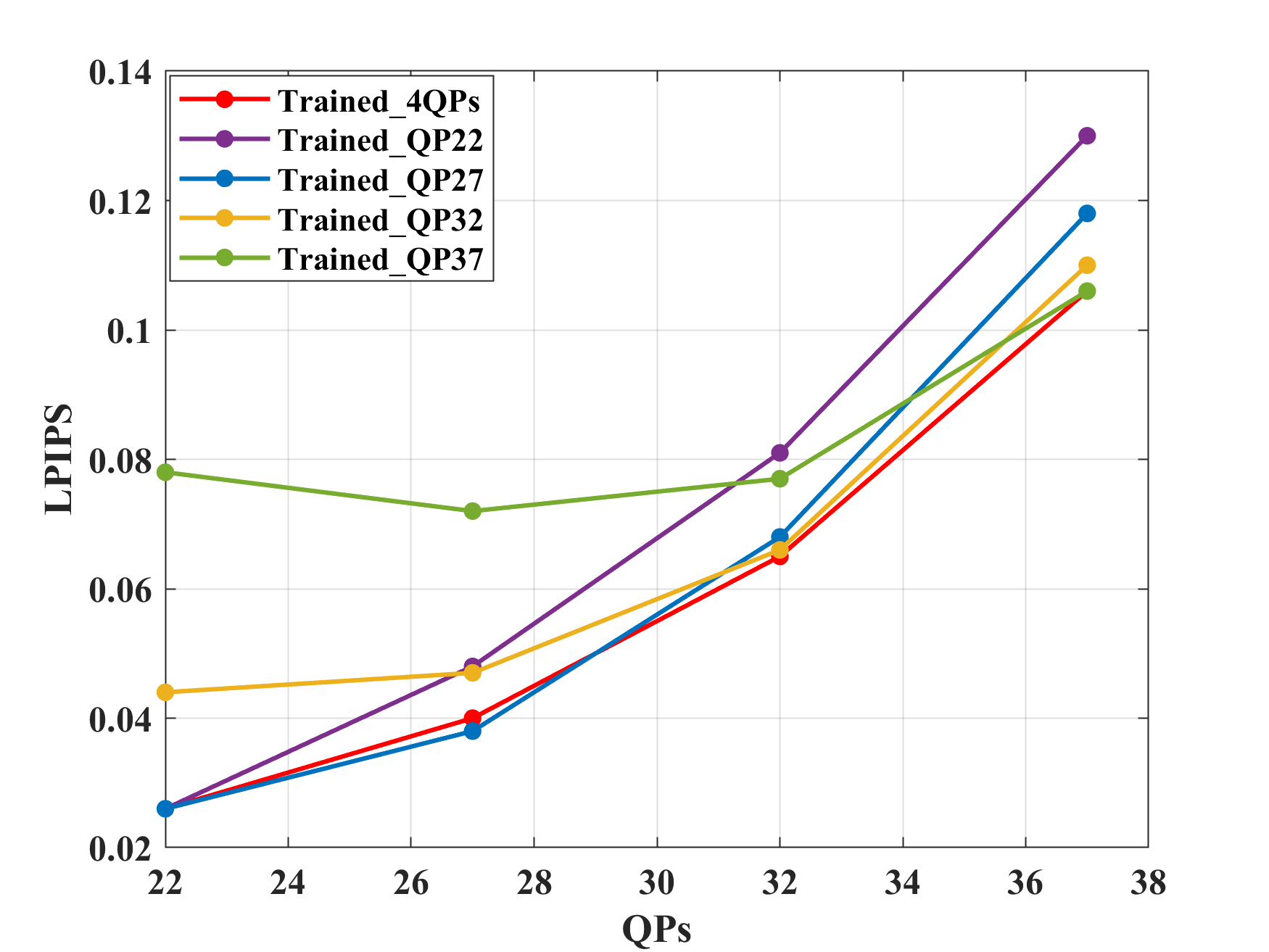}
\caption{Performance comparison of models trained at different QPs.}
\label{fig 8}
\end{figure}

\setlength\tabcolsep{1pt}
\begin{table}[t]
\caption{{Ablation Studies of QP-Conditional Adaptation Strategy Tested on JCT-VC Standard Test Sequences of H.265/HEVC at Four QPs}}
\begin{center}
\begin{tabular}{ccc}
\toprule
{Models} & {Average LPIPS} & {Number of Parameters} \\
\midrule
{with QP-conditional adaptation}     &  {0.059}      & {26329}         \\
{without QP-conditional adaptation}  &   {0.059}      & {105316} \\
\bottomrule
\end{tabular}
\end{center}
\end{table}

\setlength\tabcolsep{4.5pt}
\begin{table}[t]
\caption{{Average Performance Comparison of Different Alignment Modules on JCT-VC Standard Test Sequences of H.265/HEVC}}
\begin{center}
\begin{tabular}{ccccccc}
\toprule
        & \multicolumn{2}{c}{Deformable Convolution} & \multicolumn{2}{c}{Optical Flow} & \multicolumn{2}{c}{Attention} \\
\midrule
        & LPIPS               &DISTS               & LPIPS           & DISTS          & LPIPS         & DISTS         \\
Average &0.1149                      &0.0648                     &0.1119                &0.0568                 &\textbf{0.1059}               &\textbf{0.0565}  \\
\bottomrule
\end{tabular}
\end{center}
\end{table}

\begin{figure*}[t]
\centering
\subfloat[\textit{BasketballPass}]{\includegraphics[width=0.3\textwidth]{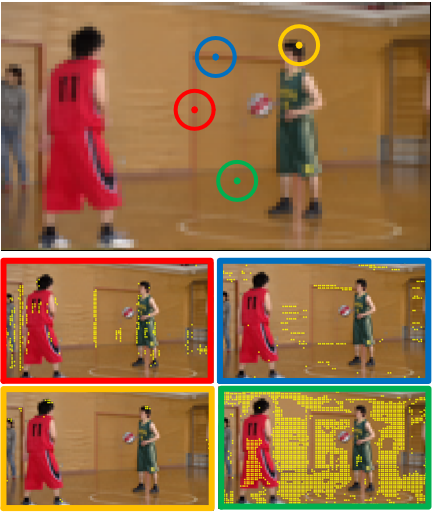}
\label{fig 9 (a)}}
\hfil
\subfloat[\textit{BQSquare}]{\includegraphics[width=0.3\textwidth]{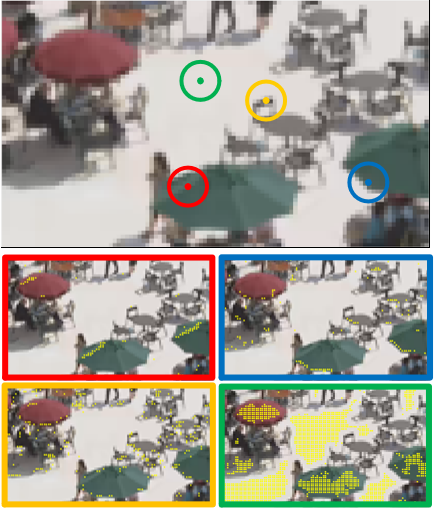}
\label{fig 9 (b)}}
\hfil
\subfloat[\textit{RaceHorses}]{\includegraphics[width=0.3\textwidth]{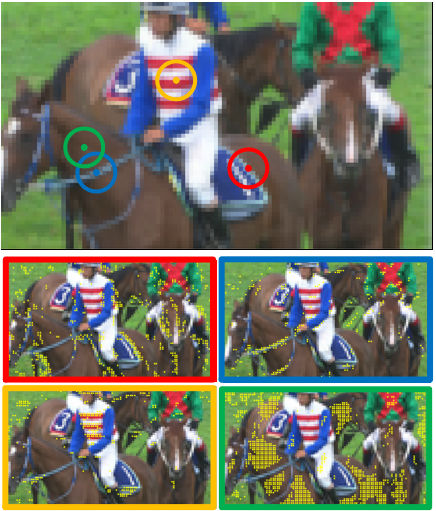}
\label{fig 9 (c)}}
\caption{Captured global temporal correlations.}
\label{fig 9}
\end{figure*}

\subsection{QP-Conditional Adaptation Evaluation}

Since the proposed PeQuENet can achieve the QP-conditional adaptation, we only train one model to enhance the perceptual quality of compressed videos at the four different QPs. The performance of this model, referred to as ``Trained\_4QPs", has been shown in TABLE I (at the last two columns). To validate the effectiveness of the QP-conditional adaptation strategy, we train the four models, named ``Trained\_QP22", ``Trained\_QP27", ``Trained\_QP32" and ``Trained\_QP37", for the four QPs. It should be noted that the training dataset for each of these models contains only samples compressed at the corresponding single QP. The performance of the four models tested on the four QPs is shown in TABLE II. {To highlight the advantage of the QP-conditional adaption strategy used in the proposed PeQuENet, the average LPIPS performance and the number of parameters of the model with or without QP-conditional adaptation are compared in TABLE III.} Besides, the the average LPIPS comparison between ``Trained\_4QPs" and the other four models is shown in Fig. 8.

Specifically, the model trained at a single QP performs the best when tested on this QP since the model can learn the characteristics of videos compressed at the corresponding QP in the training process. For instance, the model ``Trained\_QP37" (i.e., the green line in Fig. 8) shows its best performance compared with the other models (especially ``Trained\_QP22") when tested on the QP 37. This is because ``Trained\_QP37" have learned specific characteristics of videos compressed at the QP 37 after training while the other models do not have access to such characteristics. However, ``Trained\_QP37" has the worst performance when tested on the QP 22. It is not only because it has not learned the characteristics of videos compressed at the QP 22, but also because the differences of characteristics between videos compressed at the QPs 22 and 37 are significant. On the other hand, with training dataset containing samples compressed at the four QPs and encoded QP information fed into the model, ``Trained\_4QPs" achieves similar results compared to the models trained for specific QPs, which illustrates the effectiveness of the used QP-conditional adaption strategy.

\begin{figure}[t]
\centering
\includegraphics[width=0.48\textwidth]{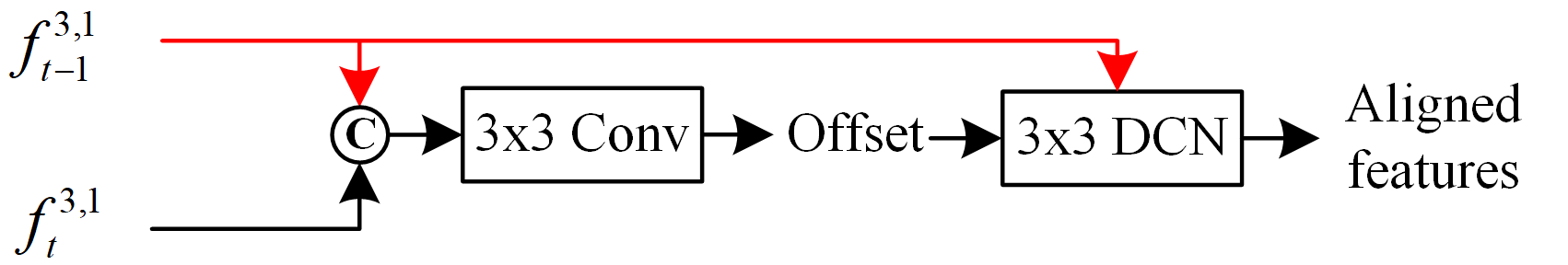}
\caption{The deformable convolution (DCN) block.}
\label{fig 10}
\end{figure}

\begin{figure}[t]
\centering
\includegraphics[width=0.48\textwidth]{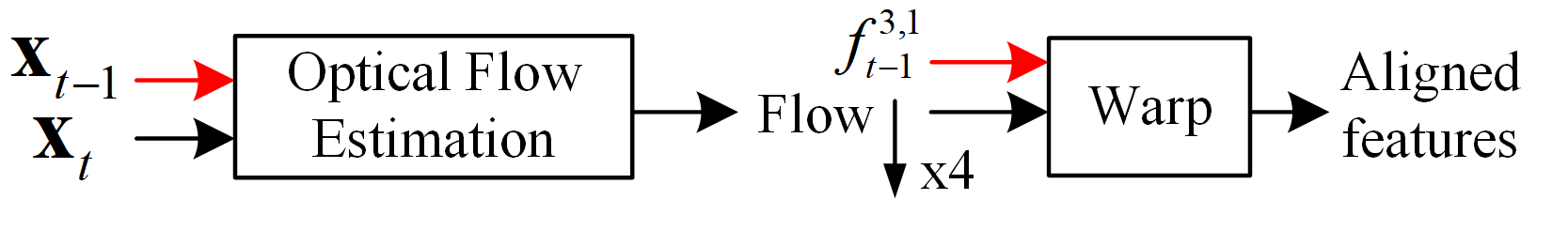}
\caption{The optical flow block. ↓x4 means downsampling by the factor 4.}
\label{fig 11}
\end{figure}

\setlength\tabcolsep{2pt}
\begin{table}[t]
\caption{{Overall Speed (frames per second, fps) on JCT-VC Standard Test Sequences of H.265/HEVC Measured on Nvidia GeForce GTX 2080 Ti GPU}}
\begin{center}
\begin{tabular}{ccccc}
\toprule
\multicolumn{2}{c}{Sequences}                                & Deformable Convolution & Optical Flow & Attention \\
\midrule
\multirow{2}{*}{\textit{Class A}} & \textit{Traffic}         &\textbf{0.48}                         &0.36               &0.36           \\
                                  & \textit{PeopleOnStreet}  &\textbf{0.47}                         &0.37               &0.36           \\
\multirow{5}{*}{\textit{Class B}} & \textit{Kimono}          &\textbf{0.94}                         &0.71               &0.76           \\
                                  & \textit{ParkScene}       &\textbf{0.94}                         &0.72               &0.76           \\
                                  & \textit{Cactus}          &\textbf{0.93}                         &0.71               &0.75           \\
                                  & \textit{BQTerrace}       &\textbf{0.94}                         &0.72               &0.76           \\
                                  & \textit{BasketballDrive} &\textbf{0.93}                         &0.71               &0.76           \\
\multirow{4}{*}{\textit{Class C}} & \textit{RaceHorses}      &\textbf{4.68}                         &3.47               &3.71           \\
                                  & \textit{BQMall}          &\textbf{4.71}                         &3.52               &3.68           \\
                                  & \textit{PartyScene}      &\textbf{4.66}                         &3.49               &3.67           \\
                                  & \textit{BasketballDrill} &\textbf{6.68}                         &3.49               &3.71           \\
\multirow{4}{*}{\textit{Class D}} & \textit{RaceHorses}      &16.01                        &10.77              &\textbf{20.00}          \\
                                  & \textit{BQSquare}        &16.28                        &11.07              &\textbf{19.82}          \\
                                  & \textit{BlowingBubbles}  &16.27                        &10.96              &\textbf{19.76}          \\
                                  & \textit{BasketballPass}  &15.64                        &10.83              &\textbf{18.80}          \\
\multirow{3}{*}{\textit{Class E}} & \textit{FourPeople}      &\textbf{2.08}                         &1.59               &1.81           \\
                                  & \textit{Johnny}          &\textbf{2.10}                         &1.59               &1.79           \\
                                  & \textit{KristenAndSara}  &\textbf{2.05}                         &1.60               &1.81           \\
\hline
\multicolumn{2}{c}{\textit{Average}}                         &5.27                         &3.70               &\textbf{5.73}           \\
\bottomrule
\end{tabular}
\end{center}
\end{table}

\begin{figure*}[t]
\centering
\subfloat[Blocking artifact in \textit{BasketballPass}]{\includegraphics[width=0.95\textwidth]{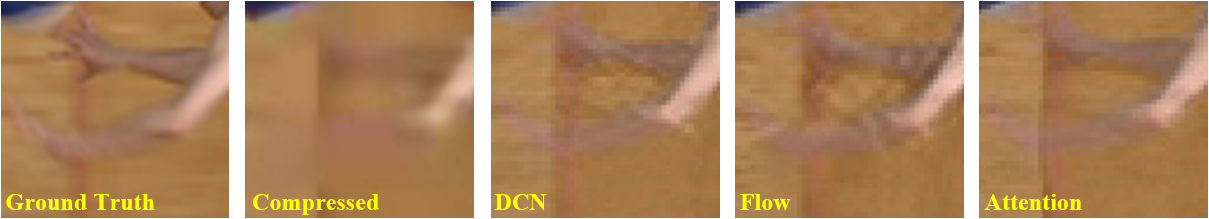}
\label{fig 12 (a)}}
\hfil
\subfloat[Ringing artifact in \textit{ParkScene}]{\includegraphics[width=0.95\textwidth]{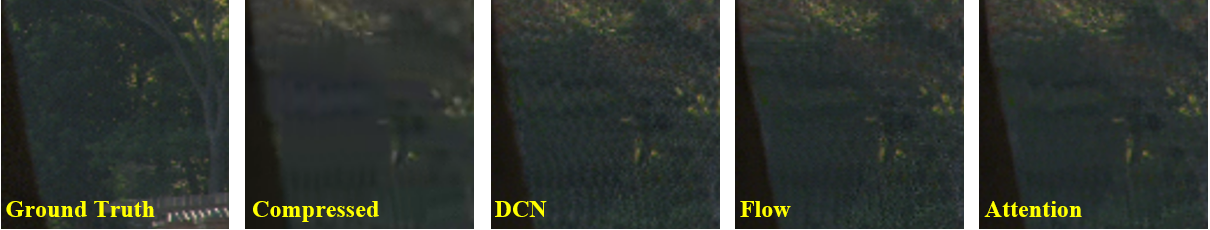}
\label{fig 12 (b)}}
\hfil
\subfloat[Mixture of various artifacts in \textit{Kimono}]{\includegraphics[width=0.95\textwidth]{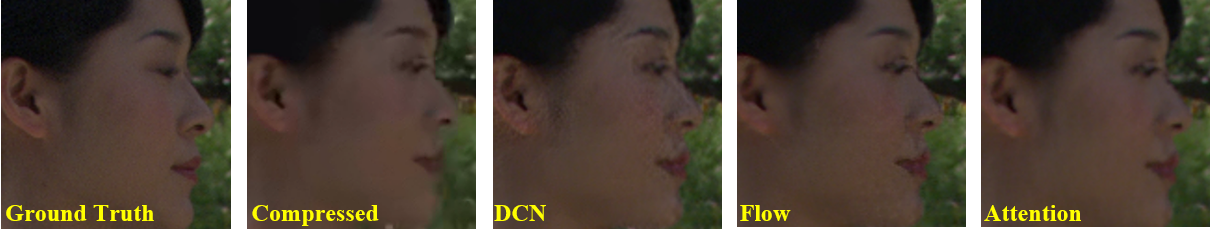}
\label{fig 12 (c)}}
\caption{Different artifacts introduced by local alignment modules.}
\label{fig 12}
\end{figure*}

\begin{figure}[]
\centering
\includegraphics[width=0.5\textwidth]{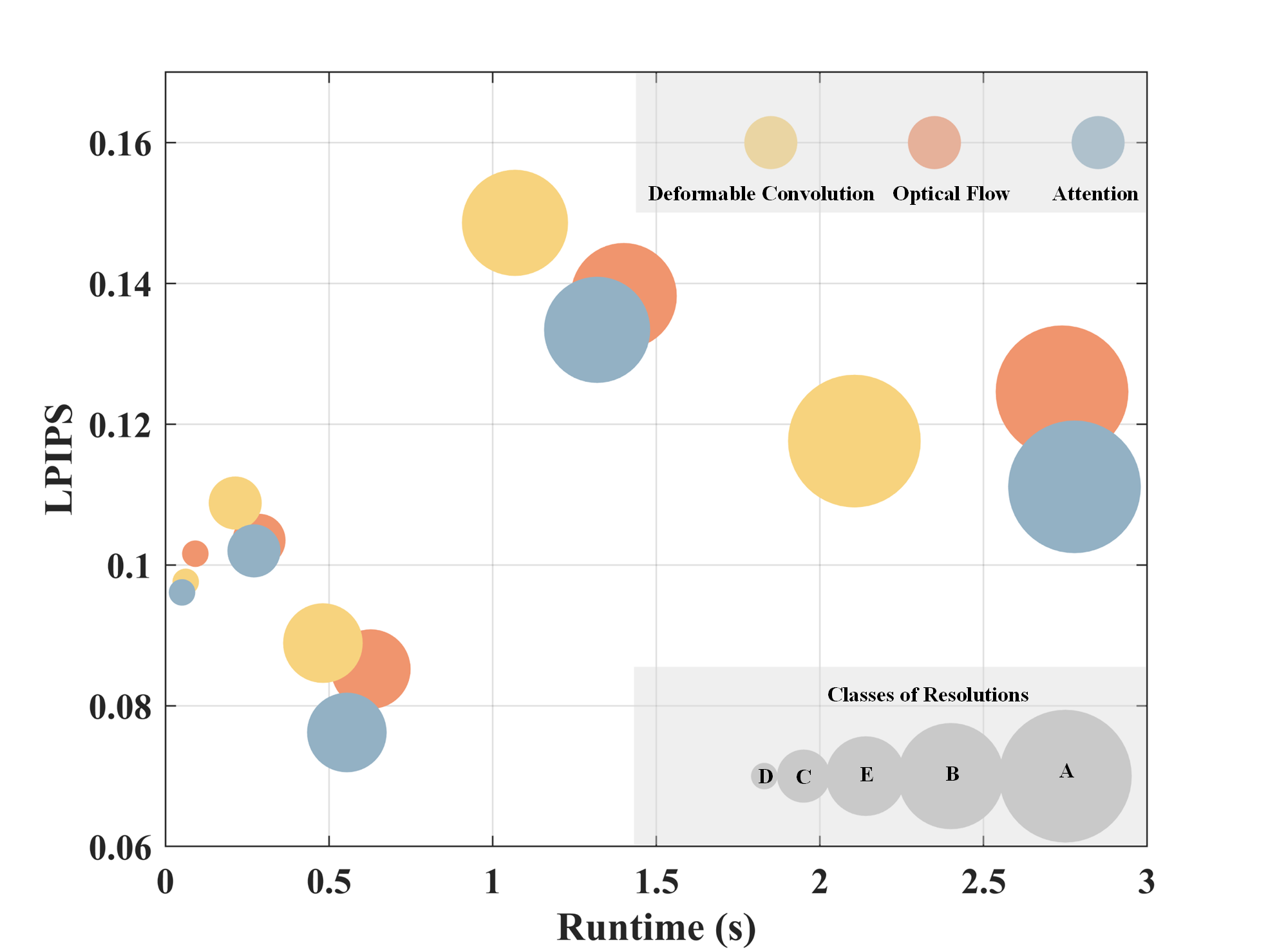}
\caption{Performance and runtime (s) comparison.}
\label{fig 13}
\end{figure}

\subsection{Visualization of Global Temporal Correlations}
The non-local attention module aims at leveraging global temporal correlations in consecutive frames to enhance the perceptual quality of compressed videos in the proposed PeQuENet. In this section, taking the leftmost attention block drawn in Fig. 2 (six attention blocks in total) as an example, we visualize its captured global temporal correlations, as shown in Fig. 9. It should be noted that, due to down-sampling operations, the resolution of the input of this attention block is a quarter of that of an input frame. To cater for the resolution, frames shown in Fig. 9 are also down-sampled in the same way.

Specifically, the frames shown in the first row are the third frame in the corresponding video sequences compressed at QP 37, and other frames are the corresponding preceding frame (i.e., the second frame). Four cases are selected in each frame to illustrate the captured global temporal correlations. For example, when the pixel is located at the vertical line on the wall (i.e., the red pixel emphasized by a red circle in Fig. 9 (a)), its correlated pixels in the preceding frame are all located at vertical lines. While the pixel is located at the horizontal line on the wall (i.e., the blue pixel emphasized by a blue circle in Fig. 9 (a)), its correlated pixels in the preceding frame are all located at the horizontal line. Moreover, these correlated pixels are distributed globally and not limited by the range of the motion, which provides more temporal information to achieve advanced and robust performance in the perceptual quality enhancement of compressed videos.

\subsection{Global Alignment vs. Local Alignment}
In this section, we compare the global alignment, i.e., attention maps, with two common local alignment operations, i.e., deformable convolutions and optical flows, to emphasize the importance of leveraging global temporal information in the perceptual quality enhancement of compressed videos. Note that the deformable convolution is applied in the STDF {and DCNGAN}, and the optical flow is applied in the MFQE 2.0 and MW-GAN. These four quality enhancement networks have been compared with our proposed PeQuENet in TABLE I and Fig. 5.

\setlength\tabcolsep{10.5pt}
\begin{table*}[t]
\caption{Ablation Studies on JCT-VC Standard Test Sequences of H.265/HEVC at QP 37 on Nvidia GeForce GTX 2080 Ti GPU}
\begin{center}
\begin{threeparttable}[b]
\begin{tabular}{cccccccccc}
\toprule
Components\tnote{*}                   & A & B & C & D & E & F & G & H & \begin{tabular}[c]{@{}c@{}}I\\ PeQuENet (Proposed)\end{tabular} \\
\midrule
Target Frame       &\checkmark   &\checkmark   &\checkmark   &\checkmark   &\checkmark   &\checkmark  &\checkmark   &\checkmark  &\checkmark                                                             \\
Preceding Frame    &   &   &\checkmark   &\checkmark   &\checkmark   &\checkmark   &\checkmark   &\checkmark   &\checkmark                                                             \\
Succeeding Frame   &   &   &   &   &   &   &   &\checkmark   &\checkmark                                                             \\
Feature Extraction &\checkmark   &\checkmark   &\checkmark   &\checkmark   &\checkmark   &\checkmark  &\checkmark   &\checkmark  &\checkmark                                                             \\
Attention Block 1  &   &   &\checkmark   &\checkmark   &\checkmark   &\checkmark   &\checkmark   &\checkmark   &\checkmark                                                             \\
Attention Block 2  &   &   &   &   &   &   &\checkmark   &\checkmark   &\checkmark                                                             \\
Attention Block 3  &   &   &   &\checkmark   &\checkmark   &\checkmark   &\checkmark   &\checkmark   &\checkmark                                                             \\
Attention Block 4  &   &   &   &   &   &   &   &\checkmark   &\checkmark                                                             \\
Attention Block 5  &   &   &   &   &\checkmark   &\checkmark   &\checkmark   &\checkmark   &\checkmark                                                             \\
Attention Block 6  &   &   &   &   &   &   &   &   &\checkmark                                                             \\
First Branch       &   &   &   &   &   &\checkmark   &\checkmark   &\checkmark   &\checkmark                                                             \\
Second Branch     &\checkmark   &\checkmark   &\checkmark   &\checkmark   &\checkmark   &\checkmark  &\checkmark   &\checkmark  &\checkmark                                                           \\
QP-conditional adaptation      &   &\checkmark   &\checkmark   &\checkmark   &\checkmark   &\checkmark  &\checkmark   &\checkmark  &\checkmark                                                             \\
\hline
LPIPS              &0.176   &0.174   &0.164   &0.162   &0.161   &0.125   &0.107   &0.107   &0.106 \\
Speed (fps)        &17.99   &16.51   &9.57   &9.40   &9.36   &7.57   &5.83   &5.76   &5.73 \\
\bottomrule
\end{tabular}
\begin{tablenotes}
\item[*] As shown in Fig. 2, attention blocks are sorted from the left to the right, represented by ``Attention Block 1", ``Attention Block 2",…and ``Attention Block 6". “First Branch” is the upper branch in the progressive decoder module, and “Second Branch” is the lower branch in the progressive decoder module. LPIPS and speed are averaged on all standard test sequences of JCT-VC. Note that one convolution layer is utilized to change channels when QP-conditional adaptation module is removed.
\end{tablenotes}
\end{threeparttable}
\end{center}
\end{table*}

\setlength\tabcolsep{18pt}
\begin{table}[t]
\caption{{Ablation Studies on JCT-VC Standard Test Sequences of H.265/HEVC at QP 37}}
\begin{center}
\begin{tabular}{cccc}
\toprule
{Loss} & {A} & {B} & \begin{tabular}[c]{@{}c@{}}{C (Proposed)}\end{tabular}\\
\midrule
${L_G}$      &  {\checkmark}      & {\checkmark}       &  {\checkmark}          \\
${L_{vgg}}$  &        &   {\checkmark}      &   {\checkmark}         \\
${L_{fm}}$   &        &        &  {\checkmark}         \\
\midrule
{LPIPS}     &   {0.227}      &  {0.108}       & {\textbf{0.106}}    \\
\bottomrule
\end{tabular}
\end{center}
\end{table}

Keeping the framework of the network unchanged, we replace attention blocks with deformable convolution blocks or optical flow blocks. For example, the leftmost attention block shown in Fig. 2 is replaced by the deformable convolution block shown in Fig. 10 or the optical flow block shown in Fig. 11. After aligned features are obtained, they are concatenated with the output of the corresponding layer in the progressive decoder module to provide temporal information for the target frame. Full structures of the networks incorporating deformable convolutions and optical flows are shown in Fig. 16 and Fig. 17 in our \textit{Supplemental Material}. It should be noted that the structure of the deformable convolution block is referred to that in the PCD module in \cite{b41}, and that optical flow estimation in the optical flow block is achieved by the pre-trained model of SPynet \cite{b42}. The average performance tested on all JCT-VC standard test sequences compressed at QP 37 is compared in TABLE IV.

Essentially, deformable convolutions and optical flows are local temporal alignment operations since the range of the temporal information they captured is determined by the range of the motion. Limited by being relatively local, deformable convolutions and optical flows penalize the robustness of the perceptual quality enhancement, which results in some artifacts, e.g., blocking and ringing, in the output videos, as shown in Fig. 12. On the other hand, the attention performs non-local temporal alignment and averaging. It has access to global temporal information, which ensures its success in the perceptual quality enhancement task.

We also compare the computational complexity of three models utilizing the three alignment modules respectively, as shown in TABLE V. The model with the attention module achieves the fastest speed (almost twice as fast as the model with the optical flow module) on sequences in \textit{Class D}, which illustrates the highest efficiency of attention maps for sequences in low resolution. While deformable convolutions have advantages in processing sequences in high resolution with a speed faster than that of attention maps. In summary, the speed of the model with the attention module is the fastest averaged on all standard test sequences. It should be noted that we split frames into parts and process each part independently if sequences in high resolution cannot be fed into the model with the attention module at once due to the limitation of memory. After enhancement, we merge enhanced parts into a whole frame. The computational complexity shown in the last column in TABLE V is the complexity of the whole process. Fortunately, there are no artifacts caused by splitting and merging operations.

For better illustration, the relations between the performance (LPIPS) and the runtime (second) of the models with the three alignment modules are shown in Fig. 13. The model with the attention module achieves the advanced performance while maintaining efficiency to some extent.

\subsection{Performance-Speed Tradeoff}
\subsubsection{Downsampled Attention Maps}
Although the attention mechanism helps the proposed PeQuENet achieve advanced performance, the attention map which is in $hw \times hw$ resolution (i.e., complete attention maps), as shown in Fig. 3, requires a great number of multiplication operations and large memory to store especially for high-resolution sequences. To prompt the proposed PeQuENet to be more efficient, the inputs of attention blocks are downsampled by factors 2, 4 and 6, to obtain attention maps in reduced resolutions. In this case, the proposed PeQuENet can achieve faster speed with limited performance loss. For better comparison, we also test the performance and speed when the attention mechanism is removed from the proposed PeQuENet (i.e., w/o attention maps). The average performance and speed trade-off on all JCT-VC standard test sequences at QP 37 is shown in Fig. 14.

\subsubsection{Ablation Studies of Components}

Ablation studies are performed to better understand the contribution and the complexity of each component in the proposed PeQuENet. We start from a baseline and gradually add components on it. The LPIPS performance and speed of each configuration (from A to I) are shown in TABLE VI. It is apparent that each component brings performance improvement (i.e., LPIPS ranging from 0.176 to 0.106) with additional computational complexity (i.e., speed ranging from 17.99 fps to 5.73 fps). By choosing different configurations, the proposed PeQuENet can achieve trade-off between the performance and speed, as shown in  Fig. 15.

\begin{figure}[]
\centering
\includegraphics[width=0.45\textwidth]{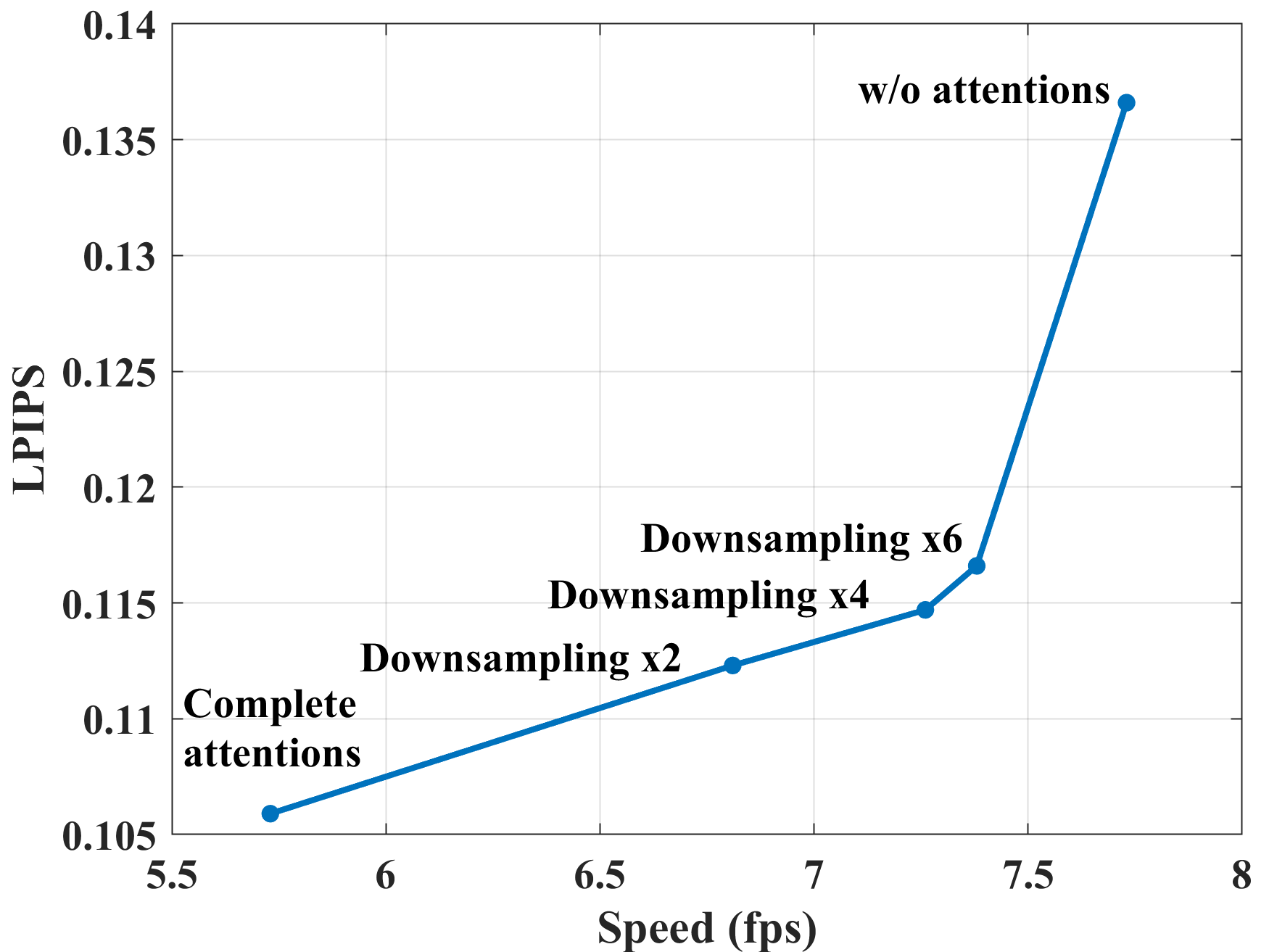}
\caption{Trade-off between performance and speed (fps) by downsampling attention maps. (Nvidia GeForce GTX 2080 Ti GPU)}
\label{fig. 14}
\end{figure}

\begin{figure}[]
\centering
\includegraphics[width=0.43\textwidth]{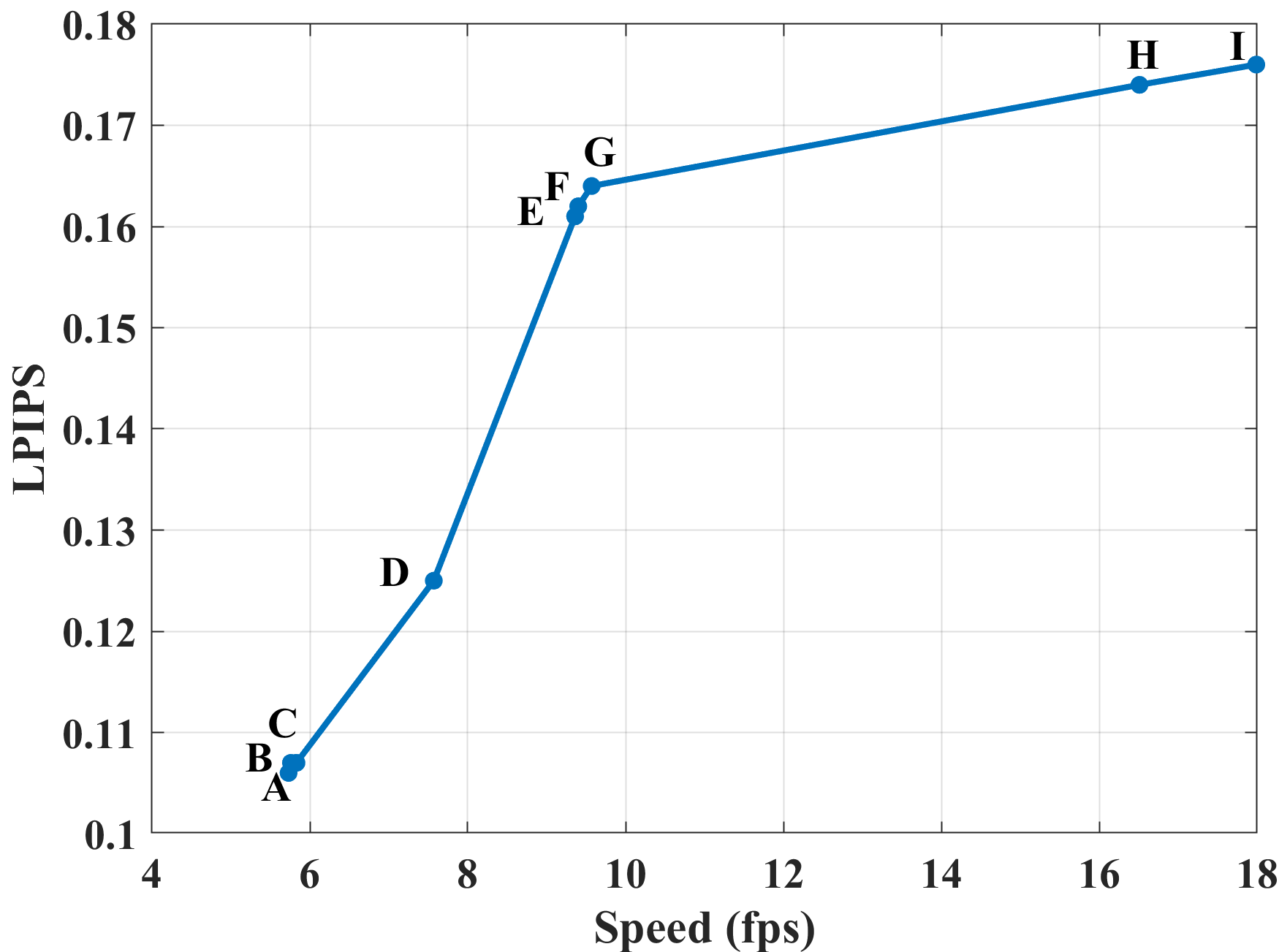}
\caption{Trade-off between performance and speed (fps) under different configurations. (Nvidia GeForce GTX 2080 Ti GPU)}
\label{fig. 15}
\end{figure}

\setlength\tabcolsep{6.5pt}
\begin{table*}[t]
\caption{{Comparison of the Number of Model Parameters}}
\begin{center}
\begin{tabular}{cccccccc}
\toprule
 & {MFQE 2.0 \cite{b9}} & {STDF \cite{b10}} & {MW-GAN \cite{b13}} & {VPE-GAN \cite{b14}} &{MPRNet \cite{b15}} &{DCNGAN \cite{DCNGAN}} &{Proposed}\\
\midrule
{Number of parameters} & {255K}     &   {365K}      &  {53839K}       & {11376K} &{333K} & {45977K} &{26329K}   \\
\bottomrule
\end{tabular}
\end{center}
\end{table*}

\setlength\tabcolsep{1.7pt}
\begin{table*}[htbp]
\caption{{Overall Performance on PSNR (dB) and SSIM of JCT-VC Standard Test Sequences of H.265/HEVC}}
\begin{center}
\begin{threeparttable}[b]
\begin{tabular}{ccccccccccccccccccc}
\toprule
{\multirow{2}{*}{{QP}}}                & \multicolumn{2}{c}{\multirow{2}{*}{{Sequences}}} & \multicolumn{2}{c}{{Compressed}}  & \multicolumn{2}{c}{{MFQE 2.0\cite{b9}}} & \multicolumn{2}{c}{{STDF\cite{b10}}} &  \multicolumn{2}{c}{{MW-GAN\cite{b13}}} & \multicolumn{2}{c}{{VPE-GAN\cite{b14}}} & \multicolumn{2}{c}{{MPRNet\cite{b15}}} &
\multicolumn{2}{c}{{DCNGAN\cite{DCNGAN}}} &\multicolumn{2}{c}{{Proposed}} \\
\multirow{20}{*}{{32}} & \multicolumn{2}{c}{}         & {PSNR}         & {SSIM}  & {PSNR}         & {SSIM}      & {PSNR}         & {SSIM}        & {PSNR}         & {SSIM}      & {PSNR}         & {SSIM}        & {PSNR}         & {SSIM}   & {PSNR}         & {SSIM}    & {PSNR}         & {SSIM}\\
\midrule
                     & \textit{{Class A}}                    & \textit{{Traffic}}   &{35.92}     &{0.9311}    &{36.52}    &{0.9378}   &{36.61}      & {0.9383} &{34.89}   & {0.8991}     &{25.42}              &{0.8079}   &{34.58}    & {0.9159}     &  {34.78}   & {0.9174}     & {35.48}    &{0.9169}               \\
                     &                            & \textit{{PeopleOnStreet}}     &{34.22}    &{0.9079}    &{34.94}    &{0.9165}   &  {35.12}    & {0.9179}  &{33.75}   &{0.8979}      &{31.55}              &{0.8641}   & {33.29}   &  {0.8956}    &  {33.03}   &  {0.8972}    &  {34.16}   &{0.8984}                \\
                     & \multirow{5}{*}{\textit{{Class B}}}   & \textit{{Kimono}}   & {37.00}   &{0.9149}    &  {37.44}  & {0.9219}  & {37.72}     &{0.9249}   &{35.73}   &{0.9110}      & {32.73}             &{0.8629}   &{35.65}    &{0.9006}     & {36.38}    & {0.9084}     & {36.66}    & {0.9061}                 \\
                     &                            & \textit{{ParkScene}}          &{34.17}    & {0.8823}   & {34.64}   &{0.8903}   & {34.76}     &{0.8928}   & {32.94}  &{0.8532}      &  {28.74}            & {0.7866}  &{33.13}    &{0.8635}      & {33.04}    &  {0.8634}    &{33.64}     &  {0.8651}                 \\
                     &                            & \textit{{Cactus}}            &{34.50}    &{0.8807}    &{34.92}    &{0.8868}   &{35.10}      & {0.8894}  &  {33.96} & {0.8569}     &{31.72}     &{0.8378}   & {33.11}   & {0.8652}    &{33.08}     &{0.8649}     & {34.07}    &{0.8672}                 \\
                     &                            & \textit{{BQTerrace}}         &{33.40}    &{0.8735}    & {33.72}   &{0.8778}   & {33.90}    & {0.8799}    &  {32.47}&{0.8467}      & {27.32}             & {0.7935}  &{32.11}    & {0.8548}     &{31.21}     &{0.8490}      &{33.12}     &{0.8609}                 \\
                     &                            & \textit{{BasketballDrive}}    &{35.38}    &{0.8842}    & {35.73}   & {0.8896}  &{35.91}      & {0.8914}  &  {34.67} &{0.8586}      &   {32.44}           &{0.8443}   &  {33.88}  &{0.8634}      &{35.62}     & {0.8951}     & {34.95}    &{0.8665}                \\
                     & \multirow{4}{*}{\textit{{Class C}}}   & \textit{{RaceHorses}}& {32.93}   &{0.8902}    & {33.28}   &{0.8964}   &{33.55}     & {0.8982} & {32.01}  &{0.8519}   & {30.54}     & {0.8371}             &{31.76}   &{0.8724}    & {30.81}  &   {0.8436} &{32.44}     &  {0.8697}                         \\
                     &                            & \textit{{BQMall}}             &{34.19}    & {0.9163}   &{34.77}    &
                     {0.9235}  &  {35.04}    & {0.9265}  &  {33.30} &{0.8901}      & {29.09}             & {0.8441}  & {32.90}   &{0.8977}      &  {32.96}   &{0.9213}      &{33.93} &{0.9399}               \\
                     &                            & \textit{{PartyScene}}         & {31.00}   &{0.8886} & {31.44}   &{0.8966}   &{31.76}      & {0.9000}  & {30.21}  & {0.8560}     & {26.39}     &{0.8116}   & {30.13}   & {0.8718}     &{26.56}     &  {0.8107}    &   {30.88}  &    {0.8818}              \\
                     &                            & \textit{{BasketballDrill}}    & {34.05}   &{0.8712}    &{34.56}    &{0.8782}   &{34.65}      & {0.8797}  &{32.98}   & {0.8421}     &    {31.60}          & {0.8261}  &{33.09}    &{0.8590}      &{32.03}     & {0.8534}     & {33.75}    &{0.8534}                \\
                     & \multirow{4}{*}{\textit{{Class D}}}&\textit{\textit{{RaceHorses}}}&  {32.19}  & {0.8669}   & {32.67}   &{0.8753}   & {32.83}     & {0.8790}  & {31.51}  &    {0.8349}  & {29.82}             &  {0.8019} & {31.13}   &  {0.8470}    &{31.94}     &{0.8801}      & {31.82}    &{0.8496}\\
                     &                            & \textit{{BQSquare}}          &{31.21}    &{0.8773}&    {31.45} &  {0.8825} &  {32.31} & {0.8872}  &{30.89}   & {0.8456}  &  {28.61}    &{0.8357}              & {30.29}  & {0.8641}   &  {24.90}    &  {0.7992}   &{31.17}      &{0.8685}                    \\
                     &                            & \textit{{BlowingBubbles}}    &{30.84}    & {0.8777}   & {31.18}   &{0.8788}   & {31.61}     & {0.8922}  &{28.82}   &{0.8353}      &    {28.35}          &{0.8241}   &{30.13}    &{0.8630}      &{27.58}     &{0.8073}      &{30.71}     &  {0.8723}              \\
                     &                            & \textit{{BasketballPass}}     & {33.44}   &  {0.8891}  &{34.09}    &{0.8993}   &{34.32}  & {0.9020}   &{32.65}   &{0.8640}   & {31.54}     & {0.8508}             & {32.45}  &{0.8699}    & {33.71}     &   {0.9220}  &{33.17}   &{0.8733}                       \\
                     & \multirow{3}{*}{\textit{{Class E}}}   & \textit{{FourPeople}} &  {37.49}  & {0.9503}   &  {38.23}  &   {0.9563}&   {38.32}   & {0.9569}  &{36.28}   &{0.9238}      &    {31.49}          & {0.8933}  & {35.67}   &{0.9353}      & {35.37}    & {0.9371}     &  {37.12}   &   {0.9391}                \\
                     &                            & \textit{{Johnny}}             & {38.85}   &{0.9464}    &   {39.41} & {0.9514}  & {39.47}     &{0.9515}   & {37.41}  &{0.9149}      &      {31.21}        &{0.8601}   &{33.47}    &   {0.9181}   &{38.21}     & {0.9430}     &   {38.11}  & {0.9313}                \\
                     &                            & \textit{{KristenAndSara}}    &  {38.60}  &  {0.9532}  & {39.29}   &{0.9582}   &{39.43}      & {0.9591}  &  {36.77} & {0.9148}     &   {32.33}           &{0.8934}   &   {35.50} & {0.9277}     & {36.78}    & {0.9581} &{37.88}      &{0.9353}               \\
                     & \multicolumn{2}{c}{\textit{{Average}}}                    & {34.41}   &  {0.9001}  &  {34.93}  &{0.9071}   & {35.12}     & {0.9093}  & {33.37}  &{0.8698}      &  {30.05}            &{0.8375}   &  {32.90}  &{0.8825}      & {32.67}    & {0.8817}     & {34.06}    & {0.8867}                 \\
\hline
{22}                   & \multicolumn{2}{c}{\textit{{Average}}}                   & {40.18}   &{0.9611}    & {40.64}   &{0.9638}   &  {40.75}    & {0.9641}  &{--}   &  {--}    & {32.23}             &  {0.8846} & {38.72}   & {0.9502}     & {33.37}    & {0.8979}     &  {39.70}   &{0.9560}                \\

{27}                   & \multicolumn{2}{c}{\textit{{Average}}}                    &{37.19}    &  {0.9370}  & {37.68}   &{0.9412}   &{37.86}      &{0.9423}   &{--}   & {--}     &{30.79}              & {0.8641}  &  {36.13}  &  {0.9258}    &{33.09}&    {0.8930}  &  {36.78}   &{0.9282}                 \\
{37}                 & \multicolumn{2}{c}{\textit{{Average}}}   & {31.71}   &{0.8472}    & {32.27}   & {0.8581}  &  {32.46}   & {0.8604}   &{31.06}   &{0.8268}   &  {29.80}    &{0.8020}              &{28.89}    & {0.8218}     &{30.21}     &   {0.8076}   &{31.36}     &  {0.8261}                   \\
\bottomrule 
\end{tabular}
\end{threeparttable}
\end{center}
\end{table*}

In theory, the proposed PeQuENet can be extended to take more temporal neighboring frames as input and add more attention blocks and decoded branches. However, considering the considerable performance gain when using two adjacent frames and high computational complexity introduced by adding more neighboring frames, we keep the temporal reference frames to two.

\setlength\tabcolsep{5pt}
\begin{table}[htbp]
\caption{{Overall Performance on LPIPS and DISTS of JCT-VC Standard Test Sequences of H.265/HEVC at Other QPs}}
\begin{center}
\begin{threeparttable}[b]
\begin{tabular}{ccccccc}
\toprule
{\multirow{2}{*}{{QP}}}                & \multicolumn{2}{c}{\multirow{2}{*}{{Sequences}}} & \multicolumn{2}{c}{{Compressed}}   &\multicolumn{2}{c}{{Proposed}} \\
\multirow{20}{*}{{24}} & \multicolumn{2}{c}{}         & {LPIPS}         & {DISTS}  & {LPIPS}         & {DISTS}      \\
\midrule
                     & \textit{{Class A}}                    & \textit{{Traffic}}   & {0.087} &{0.004}  &{0.024}  & {0.002}              \\
                     &                            & \textit{{PeopleOnStreet}}     & {0.065} & {0.005} &{0.026}  & {0.002}      \\
                     & \multirow{5}{*}{\textit{{Class B}}}   & \textit{{Kimono}}    & {0.193} &{0.015}  & {0.074} &{0.007}                    \\
                     &                            & \textit{{ParkScene}}          & {0.153} &{0.017}  & {0.051} & {0.007}    \\
                     &                            & \textit{{Cactus}}            &{0.180}  & {0.007} &{0.060}  &{0.003}   \\
                     &                            & \textit{{BQTerrace}}          & {0.135} & {0.013} &{0.051}  &{0.005}   \\
                     &                            & \textit{{BasketballDrive}}    &{0.179}  &{0.009}  & {0.065} & {0.003}      \\
                     & \multirow{4}{*}{\textit{{Class C}}}   & \textit{{RaceHorses}}& {0.072} &{0.029}  &{0.029}  & {0.014}           \\
                     &                            & \textit{{BQMall}}            &{0.060}  &{0.028}  &{0.020}  &{0.012}   \\
                     &                            & \textit{{PartyScene}}       &{0.031}  &{0.019}  &{0.011}  &  {0.006}      \\
                     &                            & \textit{{BasketballDrill}}   & {0.058} & {0.022} &{0.021}  &{0.009}    \\
                     & \multirow{4}{*}{\textit{{Class D}}}&\textit{\textit{{RaceHorses}}}& {0.036} &{0.058}  &{0.013}  &  {0.034}    \\
                     &                            & \textit{{BQSquare}}           & {0.029} &  {0.067}& {0.010} & {0.033}     \\
                     &                            & \textit{{BlowingBubbles}}     & {0.027} & {0.053} &{0.009}  & {0.026}  \\
                     &                            & \textit{{BasketballPass}}     &{0.039}  & {0.068} &{0.017}  & {0.037}      \\
                     & \multirow{3}{*}{\textit{{Class E}}}   & \textit{{FourPeople}} &{0.084}  & {0.017} & {0.032} & {0.004}        \\
                     &                            & \textit{{Johnny}}             &{0.118}  & {0.015} &{0.036}  & {0.004}  \\
                     &                           & \textit{{KristenAndSara}}    &{0.101}  &{0.018}  & {0.033} &{0.006}  \\
                     & \multicolumn{2}{c}{\textit{{Average}}}                    & {0.091} &{0.026}  &{0.032}  &  {0.012}   \\
\hline
{35}                   & \multicolumn{2}{c}{\textit{{Average}}}                    &{0.199}  &{0.079}  & {0.088} & {0.046}      \\
\bottomrule 
\end{tabular}
\end{threeparttable}
\end{center}
\end{table}

\subsection{Ablation Studies of Components in Total Loss}
{The contribution of each component in total loss to the improvement of perceptual quality is shown in TABLE VII. Optimized only under the guidance of the adversarial loss (i.e., $L_G$), the network tends to randomly generate artifacts, which penalizes the perceptual quality of compressed videos. By incorporating a perceptual loss (i.e., $L_{vgg}$), the network training becomes more stable, and the perceptual quality of compressed videos is improved significantly. This is because the generator learns to match intermediate representations extracted by the VGG model from the ground truth and the generated frames. Similarly, the feature matching loss (i.e., $L_{fm}$) helps the generator understand what the natural features extracted by the discriminator from the ground truth frames are like and promotes the generator to produce frames that look real. After adding the $L_{fm}$, we further obtain a slight performance gain (i.e., an LPIPS decrease from 0.108 to 0.106). In the proposed PeQuENet, we use the total loss composed of all of the three loss components to achieve the best performance. Although the total loss seems complex, it will not increase the complexity of the inference stage.}

\subsection{Comparison of the Number of Model Parameters}
{The comparison of the number of model parameters is shown in TABLE VIII. The models of the MFQE 2.0, MPRNet and STDF have fewer parameters due to simple network architectures. But the models of the proposed PeQuENet, VPE-GAN, DCNGAN and MW-GAN are relatively heavy. However, the proposed PeQuENet can significantly enhance the perceptual quality of compressed videos and achieve the best performance compared with other methods. Thus, the proposed PeQuENet is suitable to be applied in specific use cases such as video production or archiving where it is more important to get as high quality as possible, despite possible complexity increase.}

\subsection{Performance at Other QPs}
{We trained the proposed PeQuENet at four different QPs, i.e., QP 22, 27, 32 and 37, and tested the performance at these four QPs, as shown in TABLE I. To further verify the generalization capability of the proposed PeQuENet, we also test on the video sequences compressed at other QPs (i.e., QP 24 and 35) without fine-tuning. The performance on both LPIPS and DISTS is shown in TABLE X where ``Compressed" indicates the compressed videos (i.e., the input of the proposed PeQuENet) while ``Proposed" indicates the videos enhanced by the proposed method (i.e., the output of the proposed PeQuENet). As can be seen, the perceptual quality of compressed videos is improved significantly, which implies the strong generalization ability of the proposed method.}

\subsection{More Results Under H.264/AVC and H.266/VVC}
{We also evaluate the performance of the proposed PeQuENet on the standard test sequences of H.264/AVC and H.266/VVC and compare it with the other GAN-based video enhancement methods, i.e., the VPE-GAN, MPRNet and DCNGAN, to further study the effectiveness of our proposed method. Note that we directly use models trained with HEVC without fine-tuning. Specifically, standard test sequences are compressed with the relevant test models (i.e., JM\cite{b46} and VTM\cite{b47}) under the LDP configuration at QP 27 and 37. The performance is compared in TABLE XI and TABLE XII in our \textit{Supplemental Material}. As can be seen, the proposed PeQuENet can always achieve the best performance in terms of the average LPIPS and DISTS on the standard test sequences of both H.264/AVC and H.266/VVC compressed at both QPs.}

\subsection{Evaluation of Distortion}
{The proposed PeQuENet aims to enhance the perceptual quality of compressed videos and is trained with only perceptual losses. Thus, its performance measured by PSNR/SSIM is degraded because of the perception-distortion tradeoff \cite{b11}. We further evaluate the objective qualities of videos enhanced by the proposed PeQuENet and other compressed video enhancement methods and compare them in TABLE IX. As expected, the MFQE 2.0 and STDF improve PSNR and SSIM to different degrees while the other methods penalize both of them. Compared with the MW-GAN, VPE-GAN, MPRNet and DCNGAN, the proposed PeQuENet achieves the best PSNR/SSIM performance, which indicates the proposed PeQuENet achieves the best perception while having less influence on the distortion. However, the other methods, especially the VPE-GAN, enhance the perception with significantly penalizing the distortion.}

\section{Conclusions}
In this paper, PeQuENet is proposed to enhance the perceptual quality of compressed videos based on the non-local attention and QP-conditional adaptation modules. By fully leveraging global temporal information between consecutive frames, the proposed PeQuENet can provide enhanced frames in higher perceptual quality consistently. With QP-conditional adaptation, we are the first to successfully train one single model to enhance perceptual quality of videos compressed at various QPs, saving multiple sets of model parameters without performance loss. Experimental results have demonstrated that the proposed PeQuENet outperformed the state-of-the-art compressed video quality enhancement networks quantitatively and qualitatively.

%










\end{document}